\documentclass[10pt,letterpaper]{article}
\usepackage[T1]{fontenc}
\usepackage[margin=1in]{geometry}
\usepackage{times}
\usepackage{natbib}
\setcitestyle{authoryear,round,citesep={;},aysep={,},yysep={;}}
\usepackage{amsmath,amssymb,amsthm,booktabs,array,float,graphicx,microtype,multirow,placeins,longtable}
\usepackage{xcolor,colortbl}
\definecolor{resultgreen}{HTML}{2E8B57}
\definecolor{resultred}{HTML}{C84B46}
\usepackage{hyperref,url}
\hypersetup{
  pdftitle={CRAFT-GBM: Selecting Best Features for Gradient-Boosted Machine Learning},
  pdfauthor={Basil Kuok, Yi Zhou, Yao Xuan, Shuping Li, Ke-Wei Huang},
  hidelinks
}

\newcommand{\craft}{\textsc{CRAFT}\nobreakdash-GBM}

\newcommand{\E}{\mathbb E}

\newcommand{\ums}{UMS}
\newcommand{\codefield}[1]{\nolinkurl{#1}}
\newtheorem{theorem}{Theorem}
\newtheorem{proposition}{Proposition}
\newtheorem{corollary}{Corollary}
\floatstyle{ruled}
\newfloat{algorithm}{tbp}{loa}
\floatname{algorithm}{Algorithm}
\floatstyle{plain}

\title{CRAFT-GBM: Selecting Best Features for\\
Gradient-Boosted Machine Learning}
\author{%
Basil Kuok\textsuperscript{1,2,3} \quad
Yi Zhou\textsuperscript{2,3} \quad
Yao Xuan\textsuperscript{2,3} \quad
Shuping Li\textsuperscript{2,3} \quad
Ke-Wei Huang\textsuperscript{2,3}\\[0.6em]
\normalsize\textsuperscript{1}Massachusetts Institute of Technology (MIT)\\
\normalsize\textsuperscript{2}National University of Singapore (NUS)\\
\normalsize\textsuperscript{3}Asian Institute of Digital Finance (AIDF)\\[0.6em]
\normalsize\href{mailto:basil_k@mit.edu}{\nolinkurl{basil_k@mit.edu}} \quad
\href{mailto:zhouyi02@nus.edu.sg}{\nolinkurl{zhouyi02@nus.edu.sg}} \quad
\href{mailto:yaoxuan@nus.edu.sg}{\nolinkurl{yaoxuan@nus.edu.sg}}\\
\normalsize\href{mailto:aidf.li@nus.edu.sg}{\nolinkurl{aidf.li@nus.edu.sg}} \quad
\href{mailto:dishkw@nus.edu.sg}{\nolinkurl{dishkw@nus.edu.sg}}}
\date{}
\AddToHook{cmd/abstract/after}{\normalsize}

\begin{document}
\maketitle

\begin{abstract}
Feature subsampling is used to accelerate gradient-boosted trees, yet standard approaches fix the feature count in advance and largely ignore the split-gain evidence accumulated during training. We introduce CRAFT-GBM (Confidence-Ranked Adaptive Feature Tracking), a native LightGBM extension that uses this evidence to adapt which---and, by default, how many---features are searched each boosting round. CRAFT-GBM transforms root-split gains into bounded observations and constructs joint time-uniform bounds valid under adaptive, non-stationary training histories. In its default endogenous mode, after initialization, CRAFT-GBM retains exactly the features that cannot yet be ruled out as best, letting the selected feature-set size $K_t$ emerge from training history rather than a pre-specified fraction. With probability at least $1-\delta$, every feature with the largest expected gain is retained jointly across all rounds after initialization. When a feature fraction $f$ is specified instead, CRAFT-GBM ranks features by their interval point estimates and provides an approximate top-$K$ guarantee for feature budget $K_f$. We implement CRAFT-GBM within LightGBM while preserving its native feature-bundling representation, and evaluate it across 14 public and synthetic dataset families and four noise-augmented settings. In the fully endogenous setting, CRAFT-GBM reduces per-fit training time by 39.74\% with 1.04\% lower test loss in the geometric mean across all 18 RQ1 experiments. Under pre-specified feature fraction $f$, Exogenous CRAFT-GBM also yields aggregate training-time reductions up to 13.52 percentage points larger than random feature subsampling, with no increase in aggregate test loss relative to full-feature LightGBM. Accumulated training evidence can thus support adaptive feature search with statistical guarantees while retaining the practical advantages of native gradient boosting.

\end{abstract}

\noindent\textbf{CRAFT-GBM model source code:} \url{https://github.com/basilkuok/CRAFT-GBM}\par

\section{Introduction}
\label{sec:introduction}

Gradient-boosted decision trees (GBDTs) remain a strong choice for tabular
learning \citep{grinsztajn2022tree,mcelfresh2023tabular}, but
high-dimensional training is expensive across hyperparameter search,
cross-validation and retraining. Histogram-based boosting builds and scans
every searched feature's split statistics each round \citep{ke2017lightgbm}, although many features
are weak, redundant, or rarely competitive for the best split. Random feature
subsampling reduces this work, but fixes the fraction in advance and ignores
earlier split-gain evidence. Related work addresses ensemble-wide sparsity
\citep{ibrahim2024skinny} and feature-rank uncertainty \citep{neuhof2024ranking}.
We ask: \emph{can a boosting algorithm use its accumulated
split-gain evidence to decide which features still need to be searched and,
by default, how many?}

Historical gain rankings alone are an obvious but insufficient solution. A valid
screening rule must address four challenges:
(C1)~native split gains are unbounded and their scale drifts as the ensemble
fits; (C2)~a feature's gain is observed only when it is searched, so missing
observations are caused by earlier screening decisions; (C3)~gains are
non-stationary because they depend on the current learner; and
(C4)~screening decisions and the stopping round depend on the training
history, so guarantees must hold at every round and for all features at once.
Classical best-arm and subset identification handles adaptive sampling and
stopping under fixed arm means
\citep{evendar2006action,kalyanakrishnan2012pac}.
Racing handles adaptive sampling \citep{tiwari2022mabsplit}, while rising-bandit
identification assumes nondecreasing, concave rewards \citep{mussi2024rising}.
Our target needs no such shape assumptions.

We introduce \craft{} (Confidence-Ranked Adaptive Feature Tracking), a native
LightGBM extension that treats per-round feature screening as sequential
inference. Its design is theory-driven: each component is the construction a
formal guarantee requires under one of these challenges. For (C1), a feature
gain transformation maps each searched feature's best root-split gain to
\([0,1]\) using a scale fixed before the round, so observations are bounded
and comparable across rounds. For (C2) and (C3), each feature's
\emph{accumulated expected gain} averages its conditional expected
transformed gains over the rounds in which it was actually searched, a target
that follows the evolving learner without imputing gains for skipped rounds.
For (C4), empirical-Bernstein confidence sequences
\citep{howard2020chernoff,howard2021confidence} bound these targets jointly
over all features and rounds under history-dependent selection and stopping,
without independence or stationarity assumptions (Theorem~\ref{thm:coverage}). The
empirical-Bernstein construction uses squared prediction errors to control
deviations from conditional expected feature gains, with confidence interval widths
adapting to both native LightGBM's GOSS sampling variance and the accuracy of the historical predictor.

In CRAFT-GBM, two screening rules turn these bounds into a search policy. In the default
\emph{Endogenous} mode, after a 20-round full-feature warm-up, \craft{}
searches exactly the features that the joint intervals still allow to be best
(Proposition~\ref{prop:sharp}). The feature budget \(K_t\) is therefore set by the
evidence: a feature stays in while it is uncertain, is skipped once it is
confidently dominated, and can re-enter later. With probability at least
\(1-\delta\), jointly over all rounds, no feature with the largest expected
gain is dropped and every searched feature has a bounded gap to the best
(Theorem~\ref{thm:mcb}); the set-level guarantee holds for the random budget
\(K_t\) without an additional union bound (Corollary~\ref{cor:adaptive-top}).
When the user specifies a feature fraction $f$, the \emph{Exogenous} mode searches the corresponding
\(K_f\) features with the largest point estimates and bounds their gap to the
best size-\(K\) set (Theorem~\ref{thm:exogenous-k}). 

\craft{} runs inside LightGBM's training loop and uses its Exclusive Feature
Bundling layout unchanged. Both modes are compared with LightGBM-Fixed's
random subsampling and the native samplers of XGBoost and CatBoost, using
LightGBM-Full (no feature selection) as the baseline comparison. RQ2 additionally includes two ablations:
greedy mean-gain ranking (evidence without uncertainty) and exposure
balancing (exploration without evidence).
Across all 18 RQ1 experiments, Endogenous \craft{} reduces aggregate training time by
39.7\% relative to the baseline LightGBM-Full, and its aggregate test loss is 1.0\%
lower. At feature fraction \(f=0.10\), Exogenous \craft{} reduces aggregate
training time by 62.7\% with essentially unchanged test loss ($-0.04\%$).
The two ablations are about as fast, reducing training time by 63.8\% and
64.7\%, but they raise aggregate test loss by 12.7\% (greedy mean-gain
ranking) and 5.7\% (exposure balancing).
Random subsampling by LightGBM-Fixed reduces aggregate training time by 30.9\% in RQ1 and
49.2\% at \(f=0.10\) in RQ2. All values in Figures 1 and 2 are
relative to the baseline LightGBM-Full; test loss is log loss for classification
and RMSE for regression.

Our contributions are:
\begin{enumerate}
\item \textbf{Sequential inference for adaptive feature screening.}
We formulate feature screening in boosting as inference on adaptively
observed, path-dependent gains, and introduce a predictable bounded-gain
construction and training-path target that make joint time-uniform
confidence sequences applicable
(Theorem~\ref{thm:coverage}).

\item \textbf{Evidence-determined feature budgets with guarantees.}
The Endogenous rule lets uncertainty determine how many features are
searched and provides simultaneous best-feature retention and near-best
guarantees (Theorem~\ref{thm:mcb}). For a user-specified budget, the
Exogenous rule provides an approximate top-\(K\) guarantee
(Theorem~\ref{thm:exogenous-k}).

\item \textbf{Native integration and empirical efficiency.}
We integrate \craft{} directly into LightGBM while preserving its native
EFB representation. Across all 18 RQ1 experiments, it reduces aggregate
training time by 39.7\% with 1.0\% lower aggregate test loss relative to
full-feature LightGBM.
\end{enumerate}

\section{Method}
\label{sec:method}

\paragraph{Setup and notation.}
Let \(p\geq1\) count the valid features after native preprocessing by LightGBM, indexed by
\([p]=\{1,\ldots,p\}\). At boosting round \(t\), \craft{} selects a set of features (columns)
\(S_t\subseteq[p]\), with \(K_t=|S_t|\), before the native GOSS sampling of rows.
Each evaluated feature contributes its best eligible root-split gain \(G_{j,m,t}\),
whether or not it wins the split. The observed root-split gains \(G_{j,m,t}\) are mapped to the bounded interval $[0, 1]$ via Feature Gain Transformation (FGT). A feature's \emph{expected gain} \(\overline\nu_{j,t-1}\) is the conditional expectation of gain averaged across boosting rounds in which the feature is selected, whereas its
\emph{empirical mean} \(\widehat\mu_{j,t-1}\) is the average of its observed values. A best feature has the
\emph{largest expected gain} \(\nu_t^\star\), not necessarily the largest observed gain in
the current round. At round $t$, confidence sequences provide confidence intervals \([L_{j,t},U_{j,t}]\) of
expected gains \(\overline\nu_{j,t-1}\) jointly across features and rounds through round $t-1$, which endogenous feature selection uses to
retain a set of possible-best features \(S_t\), whose size \(K_t\) is a random variable. When a feature fraction \(f\) is specified, Exogenous \craft{} selects the
\(K_f=\max\{1,\lfloor fp+1/2\rfloor\}\) features with the largest
\emph{interval point estimates} \(c_{j,t}\).
Algorithm~\ref{alg:craft} in Appendix~\ref{app:round-order}
gives the selection and update procedure.

\subsection{Feature Gain Transformation (FGT)}
\label{sec:fgt}

Let \(M\geq1\) be the fixed number of native output trees in one boosting
round. For selected feature \(j\), output component \(m\), and round
\(t\), define
\begin{equation}
G_{j,m,t}
=
\max\!\left(
\{0\}\cup
\left\{
\Delta_{m,t}^{\mathrm{root}}(j,\theta):
\theta\in\Theta_{j,m,t}^{\mathrm{root}}
\right\}
\right).
\label{eq:feature-gain}
\end{equation}
Here \(\Theta_{j,m,t}^{\mathrm{root}}\) is the native eligible candidate set \(\Theta_{j,m,t}^{\mathrm{root}}\)
for feature \(j\), and \(\Delta\) is LightGBM's net eligible split gain \(\Delta_{m,t}^{\mathrm{root}}(j,\theta)\)
calculated by its native tree algorithm
\citep{chen2016xgboost,ke2017lightgbm}. The union with zero covers a selected
feature with no positive eligible candidate. Hence, given a feature,
Equation~\eqref{eq:feature-gain} gives the feature's best root
opportunity \(G_{j,m,t}\) in a boosting round \(t\), evaluated at the current learner state,
row sample, and native candidate eligibility criteria without requiring that feature
to win against other features in that round.

For output component \(m\), initialize \(s_{m,1}=\bot\). Thereafter, let the scale
\(s_{m,t}\in(0,\infty)\cup\{\bot\}\) be fixed before round \(t\), and set
\begin{equation}
\phi(g;s)=
\begin{cases}
g/(g+s),&s>0,\\
\mathbf 1\{g>0\},&s=\bot,
\end{cases}
\qquad
R_{j,t}=\frac1M\sum_{m=1}^{M}\phi(G_{j,m,t};s_{m,t}).
\label{eq:fgt}
\end{equation}
Then \(R_{j,t}\in[0,1]\). For \(s>0\), the transformation \(\phi(g;s)\) is strictly increasing,
and \(\phi(ag;as)=\phi(g;s)\) for every \(a>0\). After an accepted round, the next scale \(s_{m,t+1}\) for output \(m\) is the median of
\[
\{G_{j,m,t}:j\in S_t,\ G_{j,m,t}>0\},
\]
using the midpoint for an even number of values. This multiset contributes at
most one value per selected feature. If it is empty, the previous scale \(s_{m,t}\) is
retained. Current gains cannot affect their same-round transformation \(\phi(G_{j,m,t};s_{m,t})\).

\subsection{Uniform Moment Statistics (UMS)}
\label{sec:history}

Uniform Moment Statistics (\ums{}) maintain an equal-weight transformed-gain
sum \(A_{j,t}\), an observation count \(N_{j,t}\), and the squared-error sum \(V_{j,t}\) used below for each
feature over its selected rounds. Let
\[
I_{j,t}=\mathbf 1\{j\in S_t\},
\qquad
Y_{j,t}=
\begin{cases}
R_{j,t},&j\in S_t,\\
0,&j\notin S_t.
\end{cases}
\]
When a feature is omitted from boosting round \(t\), the assigned zero is for
bookkeeping, not to indicate that the counterfactual gain is zero. Then, starting from \(A_{j,1}=N_{j,1}=0\), for \(t\geq2\) define
\begin{equation}
A_{j,t}=A_{j,t-1}+Y_{j,t},
\qquad
N_{j,t}=N_{j,t-1}+I_{j,t}.
\label{eq:history-an}
\end{equation}
Equivalently,
\[
A_{j,t}=\sum_{s=2}^{t}Y_{j,s},
\qquad
N_{j,t}=\sum_{s=2}^{t}I_{j,s}.
\]
For \(N_{j,t}>0\), define the empirical mean \(\widehat\mu_{j,t}\) over feature \(j\)'s selected rounds as
\[
\widehat\mu_{j,t}=\frac{A_{j,t}}{N_{j,t}}.
\]

Let \((\mathcal F_t)_{t\geq1}\) describe the \emph{training history} \(\mathcal F_t\) through each accepted round.
\(\mathcal F_{t-1}\) contains the fixed data and configuration, the fixed
feature mapping and native EFB representation, every committed learner and
\craft{} update through \(t-1\), the scale \(s_{m,t}\) for round \(t\), and all earlier GOSS
draws before accepted round
\(t\). Selection is completed before the current-round GOSS sampling. Hence, \(I_{j,t}\)
is \(\mathcal F_{t-1}\)-measurable. The recorded value \(Y_{j,t}\) is
\(\mathcal F_t\)-measurable. In particular, \(\mathcal F_1\) contains the data, configuration, feature map,
and native EFB representation fixed before training, together with the
committed round-1 learner \(F_1\) and the scales \(s_{m,2}\) for round 2. It describes the history \(\mathcal F_1\)
after round 1 and before round 2.

Define the conditional gain \(\nu_{j,t}\) and, for \(N_{j,t}>0\), its accumulated average \(\overline\nu_{j,t}\):
\begin{equation}
\nu_{j,t}=\E[Y_{j,t}\mid\mathcal F_{t-1}],\qquad
\overline\nu_{j,t}
=
\frac{\sum_{s=2}^{t}\nu_{j,s}}{N_{j,t}}.
\label{eq:accumulated-target}
\end{equation}
We call \(\overline\nu_{j,t}\) the \emph{expected gain} \(\overline\nu_{j,t}\), while
\(\widehat\mu_{j,t}\) is its empirical estimator \(\widehat\mu_{j,t}\). The target \(\overline\nu_{j,t}\) is the expected
transformed root gain \(\overline\nu_{j,t}\), conditional on the preceding training history \(\mathcal F_{s-1}\),
averaged over the rounds in which that feature was evaluated.
If \(I_{j,s}=0\), then
\(Y_{j,s}=0\) and consequently \(\nu_{j,s}=0\), so the target averages
conditional transformed root gains \(\nu_{j,s}\) only over that feature's realized selected
rounds. Different features may therefore be evaluated in different learner
states. Predictability and \(0\leq Y_{j,t}\leq I_{j,t}\) imply
\(0\leq\overline\nu_{j,t}\leq1\). This target need not be stationary: it is an
average of the conditional expectations over the accumulated rounds in which the feature was selected,
not an instantaneous or recent-gain target.

\subsection{Joint Time-Uniform Bounds}
\label{sec:confidence}

The confidence bounds \([L_{j,t},U_{j,t}]\) adapt to how well past gains predict subsequent gains.
Before observing round \(t\), form the smoothed historical prediction
\(a_{j,t}=(1/2+A_{j,t-1})/(1+N_{j,t-1})\). Accumulate its squared errors:
\begin{equation}
V_{j,u}=\sum_{t=2}^{u}(Y_{j,t}-I_{j,t}a_{j,t})^2.
\label{eq:eb-variance}
\end{equation}
Only selected features contribute, since \(Y_{j,t}=I_{j,t}=0\) for an omitted feature.
Crucially, \(a_{j,t}\) is computed
\emph{before} the current gain updates \(A\) and \(N\). This error sum \(V_{j,u}\) is
observable and does not assume stationary gains or estimate an unobserved
gain for an omitted feature.

Set the per-tail error level
\(\alpha=\delta/(2p)\). Freeze a positive mixture scale \(\rho\) before
round 1 using a prespecified reference count \(n_{\mathrm{ref}}\) (default \(n_{\mathrm{ref}}=300\)), independently
of the stopping rule or any maximum-round ceiling
(Appendix~\ref{app:eb-boundary}). Let \(u_\rho(v;\alpha)\) be the
gamma-exponential mixture boundary \(u_\rho(v;\alpha)\) defined in that appendix. The radius \(r_{j,u}\) is
\begin{equation}
r_{j,u}=\frac{u_\rho(V_{j,u};\delta/(2p))}{N_{j,u}},
\qquad N_{j,u}>0.
\label{eq:boundary}
\end{equation}
This is an empirical-Bernstein confidence sequence
\citep[Theorem~4 and Proposition~9]{howard2021confidence}: smaller accumulated
prediction errors \(V_{j,u}\) give smaller radii \(r_{j,u}\) at the same observation count \(N_{j,u}\). It does
not require replacing a worst-case variance by a sample variance in an
otherwise unchanged formula.

\begin{theorem}[Joint time-uniform bounds]
\label{thm:coverage}
Condition on \(\mathcal F_1\), including the native feature map, native
EFB representation, realized finite \(p\), and the positive
\(\rho\) frozen before round 1. For every \(t\geq2\), suppose
\(I_{j,t}\) is \(\mathcal F_{t-1}\)-measurable,
\(Y_{j,t}\) is \(\mathcal F_t\)-measurable, and
\(0\leq Y_{j,t}\leq I_{j,t}\). Use the predictable predictions \(a_{j,t}\) and error
sums \(V_{j,u}\) in Equation~\eqref{eq:eb-variance}, and the radii \(r_{j,u}\) in
Equation~\eqref{eq:boundary}.
Almost surely,
\[
\Pr\!\left(
\left.
\forall j\leq p,\ \forall u\geq2:
N_{j,u}>0\Longrightarrow
\left|\widehat\mu_{j,u}-\overline\nu_{j,u}\right|
\leq r_{j,u}
\right|\mathcal F_1
\right)\geq1-\delta.
\]
\end{theorem}

For each feature, two nonnegative mixture supermartingales control the upper
and lower errors over all rounds through Ville's inequality. A union bound
over these \(2p\) events explains \(\alpha=\delta/(2p)\), and fixing the mixture
scale \(\rho\) consumes no additional error probability. The theorem therefore covers
every boosting round, any history-adapted stopping endpoint, and
valid continuation beyond \(n_{\mathrm{ref}}\), on the condition that the fixed feature universe \([p]\),
bounded transformation, and other stated invariants persist.
No independence is required among features, native EFB bundles, GOSS
sampling draws, or multiclass output trees.

\subsection{Endogenous Feature Selection}
\label{sec:mcb}

Endogenous feature selection can begin at any boosting round
\(t^\star\geq3\). Preceding rounds \(1\leq t<t^\star\) use all features
\(S_t=[p]\): round 1 initializes the transformation scales \(s_{m,2}\), and
rounds \(2\leq t<t^\star\) accumulate confidence statistics
\((A_{j,t},N_{j,t},V_{j,t})\). The theorems below hold for any such starting
round \(t^\star\geq3\), and the Endogenous CRAFT-GBM uses \(t^\star=21\) by default.

At the beginning of round \(t\),
use only statistics through \(t-1\). Define
the bounded interval \([L_{j,t},U_{j,t}]\)
\begin{equation}
(L_{j,t},U_{j,t})=
\begin{cases}
\bigl(\max\{0,\widehat\mu_{j,t-1}-r_{j,t-1}\},
      \min\{1,\widehat\mu_{j,t-1}+r_{j,t-1}\}\bigr),
      &N_{j,t-1}>0,\\
(0,1),&N_{j,t-1}=0.
\end{cases}
\label{eq:bounded-interval}
\end{equation}
Intersecting the raw interval \([\widehat\mu_{j,t-1}-r_{j,t-1},\widehat\mu_{j,t-1}+r_{j,t-1}]\) with the known range \([0,1]\) preserves coverage. An unseen
feature receives the complete parameter range. For endogenous rounds \(t\geq21\), let
\begin{equation}
C_t=\max_{1\leq c\leq p}L_{c,t},\qquad
S_t=\{j:1\leq j\leq p,\ U_{j,t}\geq C_t\},\qquad
K_t=|S_t|.
\label{eq:mcb}
\end{equation}
All features are reconsidered each round, with omitted features retaining
their statistics and confidence intervals unchanged to the next round, so
re-entry is possible but not guaranteed.
At least one feature attaining \(C_t\) is retained, so
\(1\leq K_t\leq p\).
All features, and hence every best feature, are retained during initialization. Let the joint rectangular confidence region \(\mathcal C_t\) be
\[
\mathcal C_t=\prod_{j=1}^{p}[L_{j,t},U_{j,t}].
\]

\begin{proposition}[\(S_t\) is the set of possible-best features in round \(t\)]
\label{prop:sharp}
For \(t\geq21\) and any nonempty intervals in \(\mathcal C_t\),
\[
S_t=
\left\{j:\exists v\in\mathcal C_t\text{ such that }
v_j=\max_{1\leq c\leq p}v_c\right\}.
\]
Thus \(S_t\) is the smallest set containing every feature that can be a
maximizer for some parameter vector allowed by the rectangular region \(\mathcal C_t\).
\end{proposition}

For \(t\geq21\), define
\[
\nu_t^\star=\max_{j\in[p]}\overline\nu_{j,t-1},\qquad
\mathcal M_t^\star=\operatorname*{arg\,max}_{j\in[p]}\overline\nu_{j,t-1},
\qquad r_t^\star=\min_{j\in\mathcal M_t^\star}r_{j,t-1}.
\]

\begin{theorem}[Endogenous feature selection guarantee]
\label{thm:mcb}
Conditional on \(\mathcal F_1\), with probability at least \(1-\delta\), the
following statements hold jointly for every boosting round \(t\geq21\):
\begin{enumerate}
\item No true maximizer is omitted:
\[
\mathcal M_t^\star\subseteq S_t,
\qquad K_t\geq|\mathcal M_t^\star|.
\]
\item For every omitted feature \(j\notin S_t\), there is a retained
feature \(c\in S_t\) such that
\[
\overline\nu_{c,t-1}>\overline\nu_{j,t-1}.
\]
\item Every retained feature \(j\in S_t\) is approximately best:
\begin{equation}
0\leq\nu_t^\star-\overline\nu_{j,t-1}
\leq2r_{j,t-1}+2r_t^\star.
\label{eq:retained-bound}
\end{equation}
Consequently,
\begin{align}
0&\leq\nu_t^\star-
\frac1{K_t}\sum_{j\in S_t}\overline\nu_{j,t-1}
\nonumber\\
&\leq2r_t^\star+
\frac2{K_t}\sum_{j\in S_t}r_{j,t-1}
\leq4\max_{j\in S_t}r_{j,t-1}.
\label{eq:mcb-average}
\end{align}
\item If \(K_t=1\), the sole retained feature is the unique maximizer.
More generally, if \(\mathcal M_t^\star=\{1,\ldots,p\}\), exact recovery is automatic.
Otherwise the sufficient separation condition
\begin{equation}
\max_{j\notin\mathcal M_t^\star}U_{j,t}
<\max_{c\in\mathcal M_t^\star}L_{c,t}
\label{eq:exact-mcb}
\end{equation}
implies \(S_t=\mathcal M_t^\star\).
\end{enumerate}
\end{theorem}

Hence, the theorem guarantees selection of all features ``not ruled out as best'' in each boosting round.

\begin{corollary}[Guarantee of selecting a set of possible-best features \(S_t\)]
\label{cor:adaptive-top}
At every round covered by Theorem~\ref{thm:mcb}, let
\[
S_t^{\star,K_t}\in
\operatorname*{arg\,max}_{S\subseteq\{1,\ldots,p\},\ |S|=K_t}
\sum_{j\in S}\overline\nu_{j,t-1}.
\]
Then, jointly,
\begin{equation}
0\leq\frac1{K_t}\left(
\sum_{j\in S_t^{\star,K_t}}\overline\nu_{j,t-1}
-\sum_{j\in S_t}\overline\nu_{j,t-1}
\right)
\leq2\max_{j\in S_t}r_{j,t-1}.
\label{eq:adaptive-top}
\end{equation}
No additional union bound over possible realized values of \(K_t\) is needed.
\end{corollary}

\subsection{Exogenous Feature Selection}
\label{sec:boundaries}

Users can instead fix a feature fraction \(f\) before training.
After initialization, Exogenous CRAFT-GBM selects an approximate
best-\(K_f\) set in each round, with \(K_f\) defined below.

\paragraph{Exogenous \(K\) through \texttt{feature\_fraction}.}
A value \(f=\texttt{feature\_fraction}\in(0,1)\), fixed before training,
instead specifies
\begin{equation}
K_f=\max\left\{1,\left\lfloor fp+\frac12\right\rfloor\right\},
\label{eq:exogenous-k}
\end{equation}
using LightGBM's native feature-count rounding rule. Rounds 1 and 2 still
evaluate all \(p\) features: round 1 initializes the FGT scales \(s_{m,2}\),
and round 2 supplies the first confidence observation \(R_{j,2}\) for every feature. For
every round \(t\geq3\), define the interval point estimate \(c_{j,t}\) and half-width \(h_{j,t}\):
\[
c_{j,t}=\frac{L_{j,t}+U_{j,t}}2,
\qquad
h_{j,t}=\frac{U_{j,t}-L_{j,t}}2.
\]
\craft{} selects exactly the \(K_f\) features with the largest \(c_{j,t}\).
This confidence interval point estimate \(c_{j,t}\) is generally different from the empirical mean
\(\widehat\mu_{j,t-1}\) because bounding the endpoints \(L_{j,t},U_{j,t}\) can make the interval \([L_{j,t},U_{j,t}]\)
asymmetric around that mean \(\widehat\mu_{j,t-1}\). Equal interval point estimates \(c_{j,t}\) are resolved by smaller
observation count \(N_{j,t-1}\), followed by smaller canonical feature identifier \(\mathrm{id}(j)\).
This tie rule favors a less-observed feature but does not alter the theorem
below. Exogenous CRAFT-GBM feature selection is deterministic given the training history, so
native LightGBM's parameter \texttt{feature\_fraction\_seed} has no effect, but the
per-node parameter \texttt{feature\_fraction\_bynode} must remain one:
additional node-level sampling could remove a selected feature at the root
while its observation count \(N_{j,t}\) was still incremented.

Let \(S_t^{(f)}\) denote the resulting size-\(K_f\) set \(S_t^{(f)}\) and let
\[
S_t^{\star,K_f}\in
\operatorname*{arg\,max}_{S\subseteq\{1,\ldots,p\},\ |S|=K_f}
\sum_{j\in S}\overline\nu_{j,t-1}.
\]

\begin{theorem}[Exogenous top-\(K\) selection]
\label{thm:exogenous-k}
Under the assumptions of Theorem~\ref{thm:coverage}, conditional on
\(\mathcal F_1\), with probability at least \(1-\delta\),
jointly at every boosting round \(t\geq3\),
\begin{align}
0
&\leq
\frac1{K_f}\left(
\sum_{j\in S_t^{\star,K_f}}\overline\nu_{j,t-1}
-\sum_{j\in S_t^{(f)}}\overline\nu_{j,t-1}
\right) \nonumber\\
&\leq
\frac1{K_f}\left(
\sum_{j\in S_t^{\star,K_f}\setminus S_t^{(f)}}h_{j,t}
+\sum_{j\in S_t^{(f)}\setminus S_t^{\star,K_f}}h_{j,t}
\right)
\leq2\max_{j\leq p}h_{j,t}\leq1.
\label{eq:exogenous-k-bound}
\end{align}
\end{theorem}

For fixed \(f\), this joint guarantee bounds the selected set's average
expected-gain gap from the best size-\(K_f\) set \(S_t^{\star,K_f}\) in that run, starting at
round 3. Exogenous selection may omit a best feature even when at most
\(K_f\) remain possible-best. The bound accounts for uncertainty in both
selected and omitted features, while targets exclude unobserved omitted-round
gains. No additional error allocation over rounds or sets is needed.

\begingroup
\makeatletter
\renewcommand\section{\@startsection{section}{1}{\z@}{-2ex\@plus -1ex\@minus -.2ex}{1ex\@plus .2ex}{\normalfont\Large\bfseries}}
\renewcommand\subsection{\@startsection{subsection}{2}{\z@}{-2ex\@plus -1ex\@minus -.2ex}{1ex\@plus .2ex}{\normalfont\large\bfseries}}
\renewcommand\paragraph{\@startsection{paragraph}{4}{\z@}{1.5ex\@plus .5ex\@minus .2ex}{-1em}{\normalfont\normalsize\bfseries}}
\makeatother
\section{Experimental Evaluation}
\label{sec:evaluation}

\begin{itemize}
\setlength{\itemsep}{0pt}
\setlength{\parsep}{0pt}
\setlength{\parskip}{0pt}
  \item \textbf{RQ1:} Can Endogenous \craft{} reduce training time while maintaining test loss relative to competitor models, using the same pre-run-determined settings?
  \item \textbf{RQ2:} How does Exogenous \craft{} compare with competitor models across a set of pre-specified feature fractions, using LightGBM-Full as the baseline reference?
\end{itemize}

\subsection{Experimental Setup and Design}

\paragraph{Datasets and evaluation.}
Ten public benchmark datasets cover binary/multiclass classification and regression.
Four synthetic dataset families cover sparse inputs (Sparse 1), categorical indicators
(Sparse 2), nonlinear regression (Dense 1), and redundant-feature classification
(Dense 2), each with 2,000 features and 100,000 rows per training/test split. All competitor models share the same fixed training/test partitions for each dataset,
with subject-disjoint HAR splits and patient-disjoint CT folds.
Each experiment has five paired measurements. The test set is withheld from training and test labels are never used to select features or any other training settings. Test loss is log loss for classification (C)
and RMSE for regression (R). RQ2 uses these 14 families. Figure~\ref{fig:rq1-overview} adds HAR\_N and YearMSD\_N, which are expanded HAR and YearMSD datasets with noise. This is done by appending eight row-permuted copies of every
feature column, which expands the column size of each dataset by nine times. However, added row-permuted columns need not have zero gain and are expected to have augmented noise.

\paragraph{RQ1: endogenous feature selection.}
At the beginning of the training, for each dataset, one untimed Endogenous \craft{} pre-run is carried out to determine the total number of boosting rounds \(t\) and average feature fraction
\(\bar f=(pt)^{-1}\sum_{s=1}^{t}K_s\), which include initialization. Endogenous \craft{} is configured to stop independently at
\(\tau =\bar t\), which is after ten boosting rounds without a training-loss improvement exceeding
\(0.001L_0\) over its previous round's loss, where \(L_0\) is initial training loss. The same pre-run-determined $\bar t$ and $\bar f$ are then used for all competitor models and the baseline LightGBM-Full model for fair comparison. 
LightGBM-Full, the baseline reference, uses all features for \(\bar t\) boosting rounds.
LightGBM-Fixed also runs for a total of \(\bar t\) rounds, but randomly samples a fixed fraction $\bar f$ of features in each round.
XGBoost \citep{chen2016xgboost} and CatBoost \citep{prokhorenkova2018catboost}
receive \(\bar f,t\) through native samplers. Hence, both competitor models also match the
feature and total running budgets of Endogenous \craft{}. All models use a learning rate of 0.05 without tuning. Figure~\ref{fig:rq1-overview} compares each competitor model's training time and test loss with those of LightGBM-Full.

\paragraph{RQ2: exogenous feature selection.}
Exogenous \craft{}, LightGBM-Fixed, XGBoost, CatBoost, and two ablations use
\(f\in\{0.10,0.30,0.50,0.70,0.90\}\).
For each training dataset, a 20-trial Bayesian hyperparameter search tunes for LightGBM-Full's optimal
rounds \(T^\star\) in \([50,2000]\) and optimal learning rate \(\eta^\star\) in \([0.01,0.20]\), minimizing
loss on an 80/20 inner split.
Then, all competitor models (Exogenous CRAFT-GBM, LightGBM-Fixed, XGBoost, CatBoost) are trained with the same hyperparameter-tuned $T^\star$ and $\eta^\star$ for each dataset, across all feature fractions \(f\in\{0.10,0.30,0.50,0.70,0.90\}\). Their performance results in Figure 2 are all relative changes compared to the baseline LightGBM-Full. There are two ablation models running on the same settings.
Mean ablation greedily ranks empirical gain means \(\widehat\mu_{j,t-1}\)
(evidence without uncertainty), whereas Exposure ablation selects the
least-observed features in each boosting round (exploration without gain evidence). 

\paragraph{Shared settings and reporting.}
All LightGBM-derived models share the same native LightGBM preprocessing, EFB bundling, and GOSS sampling at depth 6 and at most 64 leaves.
\craft{} uses \(\delta=0.05\), with \(n_{\mathrm{ref}}=300\) in RQ1
and \(n_{\mathrm{ref}}=1000\) in RQ2. All models use equal CPU core counts and a standardized computational environment within each RQ.
Total training time includes launch, loading, selection, boosting, and saving, but excludes
preparation, pre-runs, tuning, and test prediction.
In both Figure~\ref{fig:rq1-overview} and Figure~\ref{fig:rq2-overview}, training time and test loss changes are reported as \(100(r-1)\%\), where \(r\) averages
\textbf{Competitor model / LightGBM-Full ratios} geometrically within and then equally across
experiments. Hence, for all competitor models, \textbf{negative values indicate improvements in training time/test loss relative to LightGBM-Full}, and the opposite holds for positive values. LightGBM-Full is not plotted on both figures because it is used as the baseline comparison in the performance metric above (its ratios = 1 trivially).

\endgroup

\begingroup
\clearpage
\subsection{RQ1: Endogenous Feature Selection}
\begin{figure}[H]
  \centering
  \includegraphics[width=5.115in]{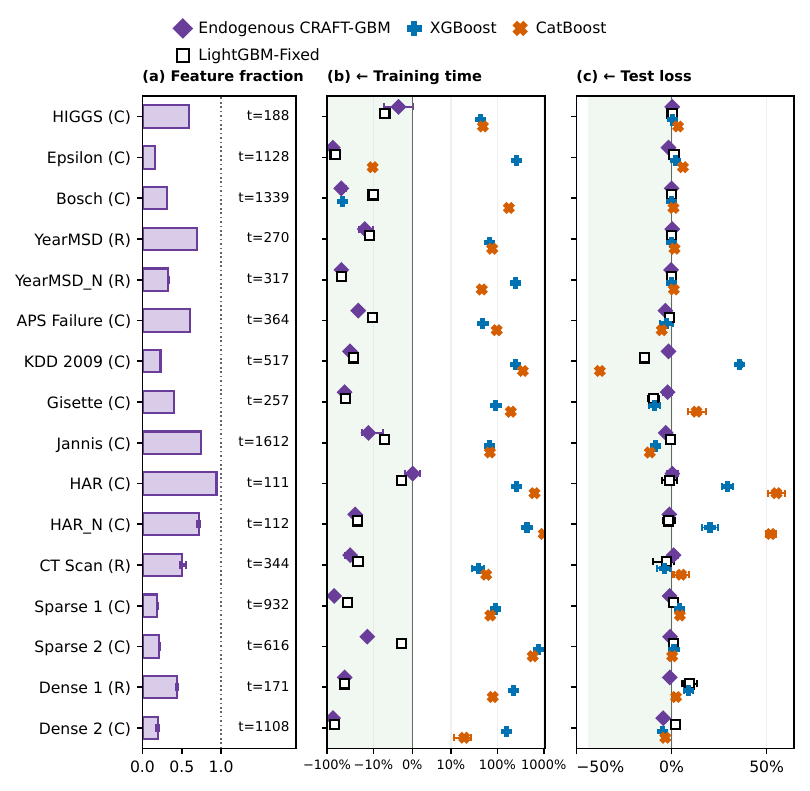}
  \par\bigskip
  \begingroup
  \begin{tabular*}{\linewidth}{@{\extracolsep{\fill}}lrrrrr@{}}
  \toprule
  & \multicolumn{2}{c}{16-experiment aggregate change} & \multicolumn{3}{c}{Wins across 16 experiments} \\
  \cmidrule(lr){2-3}\cmidrule(l){4-6}
  Model & Time (\%) & Test loss (\%) & Faster & Lower loss & Both \\
  \midrule
\textbf{Endogenous CRAFT-GBM} & \cellcolor{resultgreen!35.0000}-40.36 & \cellcolor{resultgreen!35.0000}-1.10 & \cellcolor{resultgreen!32.8125}15 & \cellcolor{resultgreen!24.0625}11 & \cellcolor{resultgreen!24.0625}11 \\
LightGBM-Fixed & \cellcolor{resultgreen!26.5835}-30.65 & \cellcolor{resultgreen!34.8156}-1.10 & \cellcolor{resultgreen!35.0000}16 & \cellcolor{resultgreen!17.5000}8 & \cellcolor{resultgreen!17.5000}8 \\
XGBoost & \cellcolor{resultred!33.8166}+135.77 & \cellcolor{resultred!35.0000}+3.95 & \cellcolor{resultgreen!2.1875}1 & \cellcolor{resultgreen!15.3125}7 & \cellcolor{resultgreen!0.0000}0 \\
CatBoost & \cellcolor{resultred!35.0000}+140.52 & \cellcolor{resultred!31.1118}+3.51 & \cellcolor{resultgreen!2.1875}1 & \cellcolor{resultgreen!8.7500}4 & \cellcolor{resultgreen!0.0000}0 \\
\bottomrule
  \end{tabular*}
  \endgroup
  \caption[Endogenous selection across 16 experiments.]{\textbf{Endogenous \craft{} reduces training time in 15 of the 16 displayed experiments.}
  Endogenous \craft{} reduces aggregate training time by 40.36\% and test loss by 1.10\%.
  LightGBM-Fixed saves 30.65\% of training time and has the same aggregate loss
  change, so endogenous \craft{} delivers a further 9.70 percentage points
  of time reduction. Endogenous \craft{} improves both outcomes in $11 / 16$ experiments, compared
  with $8 / 16$ for random LightGBM-Fixed. Across almost all datasets, XGBoost and CatBoost are slower and incur higher test loss in some datasets.
  \par\smallskip
  Time savings span 3.6--75.4\% in the
  15 faster experiments, while HAR takes 0.11\% longer. Loss falls 4.41\%
  on Dense 2 but rises 1.09\% on CT. Including both +4
  expansions leaves the overall finding similar, with 39.74\% less time
  and 1.04\% lower loss across all 18 experiments. The HAR\_N pre-run retains
  about 72\% of features, so faster training need not remove every added column.
  Together, these results support Endogenous
  \craft{} as an evidence-determined alternative to fixing a feature fraction \(f\)
  in advance, without
  implying uniformly better predictions or recovery of only useful features.}
  \label{fig:rq1-overview}
\end{figure}

\clearpage
\subsection{RQ2: Performance Across Feature Fractions}
\setlength{\intextsep}{6pt}
\setlength{\abovecaptionskip}{6pt}
\begin{figure}[H]
  \centering
  \includegraphics[width=4.51in]{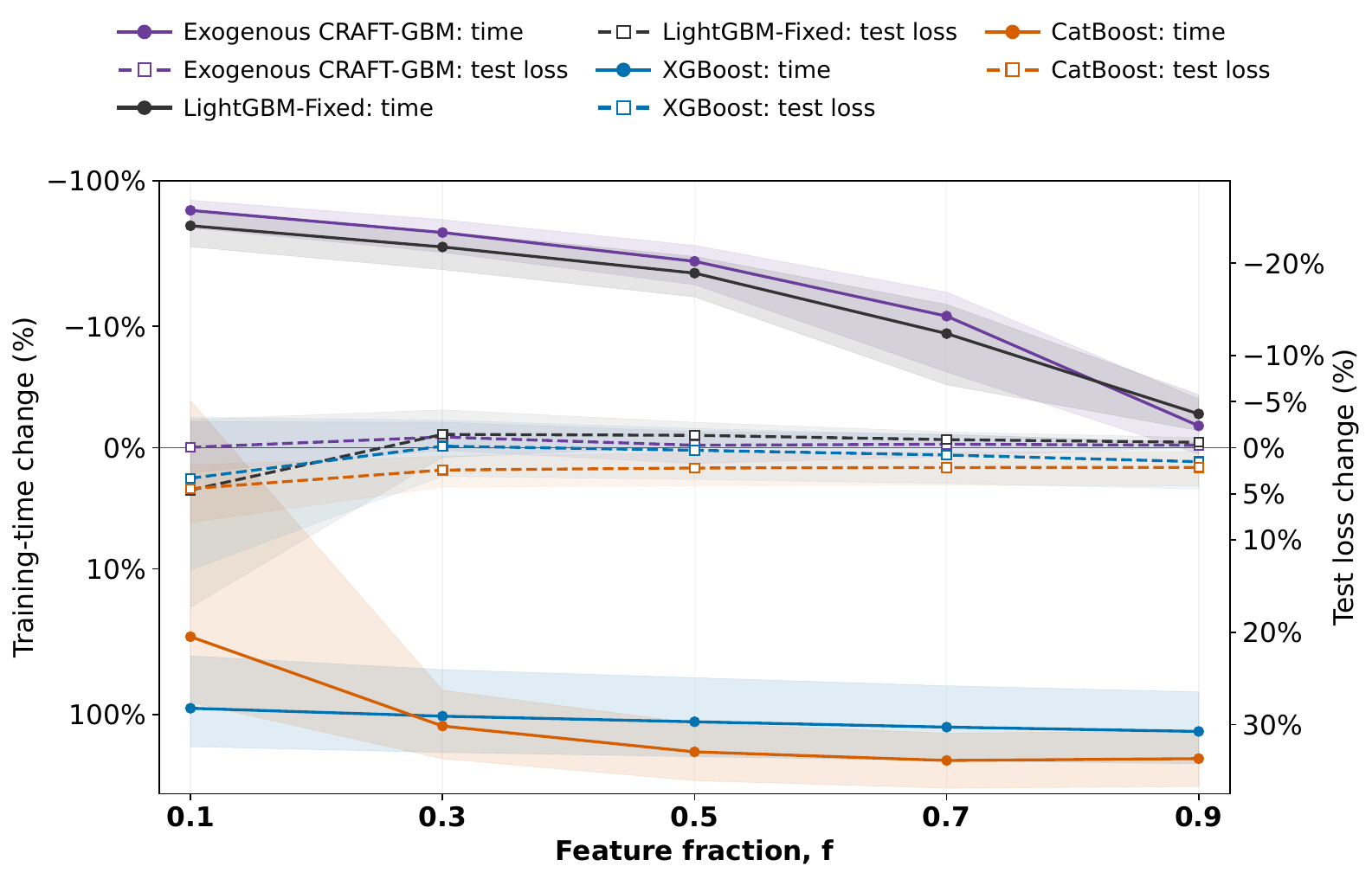}
  \par\smallskip
\begingroup
\normalsize
\setlength{\tabcolsep}{3pt}
\begin{tabular*}{\linewidth}{@{\extracolsep{\fill}}lrrrrrrrrrrrr@{}}
\toprule
\multicolumn{13}{c}{Aggregate change across 14 source families (\%)} \\
\midrule
$f$ & \multicolumn{2}{c}{\textbf{\shortstack{Exogenous\\CRAFT-GBM}}} & \multicolumn{2}{c}{\shortstack{LightGBM\\Fixed}} & \multicolumn{2}{c}{\shortstack{XGBoost}} & \multicolumn{2}{c}{\shortstack{CatBoost}} & \multicolumn{2}{c}{\shortstack{Mean\\ablation}} & \multicolumn{2}{c}{\shortstack{Exposure\\ablation}} \\
\cmidrule(lr){2-3}\cmidrule(lr){4-5}\cmidrule(lr){6-7}\cmidrule(lr){8-9}\cmidrule(lr){10-11}\cmidrule(l){12-13}
& Time & Loss & Time & Loss & Time & Loss & Time & Loss & Time & Loss & Time & Loss \\
\midrule
0.10 & \cellcolor{resultgreen!33.9345}-62.74 & \cellcolor{resultgreen!0.9722}-0.04 & \cellcolor{resultgreen!26.6219}-49.22 & \cellcolor{resultred!12.6872}+4.60 & \cellcolor{resultred!15.3176}+90.61 & \cellcolor{resultred!9.1292}+3.31 & \cellcolor{resultred!4.9362}+29.20 & \cellcolor{resultred!12.2459}+4.44 & \cellcolor{resultgreen!34.5240}-63.83 & \cellcolor{resultred!35.0000}+12.69 & \cellcolor{resultgreen!35.0000}-64.71 & \cellcolor{resultred!15.7486}+5.71 \\
0.30 & \cellcolor{resultgreen!23.9067}-44.20 & \cellcolor{resultgreen!28.1944}-1.16 & \cellcolor{resultgreen!19.0226}-35.17 & \cellcolor{resultgreen!35.0000}-1.44 & \cellcolor{resultred!17.3681}+102.74 & \cellcolor{resultgreen!4.1319}-0.17 & \cellcolor{resultred!20.2978}+120.07 & \cellcolor{resultred!6.6745}+2.42 & \cellcolor{resultgreen!24.3826}-45.08 & \cellcolor{resultred!21.3199}+7.73 & \cellcolor{resultgreen!25.4211}-47.00 & \cellcolor{resultred!5.9574}+2.16 \\
0.50 & \cellcolor{resultgreen!15.1553}-28.02 & \cellcolor{resultgreen!6.5625}-0.27 & \cellcolor{resultgreen!12.5753}-23.25 & \cellcolor{resultgreen!32.5694}-1.34 & \cellcolor{resultred!18.9724}+112.23 & \cellcolor{resultred!0.7723}+0.28 & \cellcolor{resultred!30.5337}+180.62 & \cellcolor{resultred!6.0678}+2.20 & \cellcolor{resultgreen!16.0748}-29.72 & \cellcolor{resultred!12.1355}+4.40 & \cellcolor{resultgreen!17.1620}-31.73 & \cellcolor{resultred!5.3507}+1.94 \\
0.70 & \cellcolor{resultgreen!6.3661}-11.77 & \cellcolor{resultgreen!9.2361}-0.38 & \cellcolor{resultgreen!5.0896}-9.41 & \cellcolor{resultgreen!21.3889}-0.88 & \cellcolor{resultred!20.6443}+122.12 & \cellcolor{resultred!2.1789}+0.79 & \cellcolor{resultred!35.0000}+207.04 & \cellcolor{resultred!5.9299}+2.15 & \cellcolor{resultgreen!7.1341}-13.19 & \cellcolor{resultred!9.7360}+3.53 & \cellcolor{resultgreen!7.4749}-13.82 & \cellcolor{resultgreen!24.0625}-0.99 \\
0.90 & \cellcolor{resultgreen!0.9790}-1.81 & \cellcolor{resultgreen!6.3194}-0.26 & \cellcolor{resultgreen!1.5090}-2.79 & \cellcolor{resultgreen!14.3403}-0.59 & \cellcolor{resultred!22.1015}+130.74 & \cellcolor{resultred!4.1923}+1.52 & \cellcolor{resultred!33.9823}+201.02 & \cellcolor{resultred!5.9023}+2.14 & \cellcolor{resultgreen!1.9634}-3.63 & \cellcolor{resultred!1.1308}+0.41 & \cellcolor{resultgreen!2.3690}-4.38 & \cellcolor{resultgreen!9.9653}-0.41 \\
\bottomrule
\end{tabular*}
\endgroup
  \caption[Performance across feature fractions.]{\textbf{At \(f=0.10\), Exogenous \craft{} substantially reduces training time with essentially unchanged aggregate test loss.}
  Shaded bands show pointwise 95\% dataset-resampling intervals, reflecting
  sensitivity to the composition of the 14 dataset families.
  At \(f=0.10\), Exogenous \craft{} reduces aggregate training time by 62.74\% with
  essentially unchanged test loss (\(-0.04\%\)). Random subsampling by LightGBM-Fixed saves 49.22\%
  time but raises loss by 4.60\%. Thus \craft{} delivers 13.52 percentage points
  more time reduction without the aggregate loss increase seen with LightGBM-Fixed. Greedy mean-gain ranking (Mean ablation) and exposure balancing of features (Exposure ablation)
  save the same training time
  (63.83\% and 64.71\%), but incur greater test loss especially at smaller feature fractions (12.69\% and 5.71\% more test loss at $f=0.10$).
  This contrast is consistent with the expected benefit of combining gain evidence with
  uncertainty in \craft{}, especially when the feature budget \(K_f\) is restrictive.
  \par\smallskip
  Exogenous \craft{} also has the largest training-time reduction
  among non-ablated competitor models---LightGBM-Fixed, XGBoost, and CatBoost---dominating at
  \(f\leq0.70\), while its negative test loss changes indicate lower test loss than the baseline LightGBM-Full's across all feature fractions. As expected, the advantage gained by all competitor models narrows as feature fraction increases.}
  \label{fig:rq2-overview}
\end{figure}

\endgroup

\section{Conclusion}

\craft{} turns accumulated split-gain evidence into statistically guided
feature search, deciding which features to search and, by default, how many.
Joint confidence bounds support possible-best retention and approximate
top-\(K\) selection. Relative to LightGBM-Full, Endogenous \craft{} reduces
aggregate training time by 39.74\% with 1.04\% lower test loss across all 18
RQ1 experiments. Exogenous \craft{} saves up to 62.74\% of training time at \(f=0.10\) with
essentially unchanged test loss, demonstrating efficient feature
search even under restrictive feature budgets.

\clearpage
\subsection*{Reproducibility Statement}

The appendices provide full proofs, model parameters, dataset specifications,
and complete numerical results. The supplementary source package (attached as the CRAFT-GBM\_Supplementary\_Source\_Code.zip file in the submission)
includes model implementations, data acquisition and generation scripts,
fixed experimental settings, and benchmark and analysis code. The CRAFT-GBM
model source is also available at \url{https://github.com/basilkuok/CRAFT-GBM}.
Dependency versions and execution instructions are provided to support reproduction of
the experiments. While we expect the relative performance between competitor models to be fully reproduced, the absolute training times may vary depending on the hardware and execution
environment.
\clearpage
\bibliography{arXiv_2027_November}
\bibliographystyle{plainnat}

\clearpage
\appendix

\section{Properties of Feature Gain Transformation (FGT)}
\label{app:fgt}

\begin{proof}
The claims in Section~\ref{sec:fgt} follow directly from the definition in
Equation~\eqref{eq:fgt}.

\paragraph{Boundedness.}
For \(g\geq0\) and \(s>0\),
\[
0\leq \frac{g}{g+s}<1.
\]
When \(s=\bot\), \(\phi(g;\bot)=\mathbf 1\{g>0\}\in\{0,1\}\). Hence every
summand in \(R_{j,t}\) lies in \([0,1]\), and averaging over the \(M\)
output trees gives \(R_{j,t}\in[0,1]\).

\paragraph{Monotonicity and positive-rescaling invariance.}
For a finite positive scale \(s\),
\[
\frac{\partial}{\partial g}\frac{g}{g+s}
=\frac{s}{(g+s)^2}>0,
\]
so the transformation \(\phi(g;s)\) is strictly increasing in \(g\). For every \(a>0\),
\[
\phi(ag;as)=\frac{ag}{ag+as}=\frac{g}{g+s}=\phi(g;s).
\]
Thus a common positive rescaling of the gain \(g\) and its scale \(s\) leaves the
transformed value \(\phi(g;s)\) unchanged.

\paragraph{Predictability of the scale.}
By construction, \(s_{m,t}\) is committed before accepted round \(t\) and is
therefore \(\mathcal F_{t-1}\)-measurable. The current gains
\(\{G_{j,m,t}:j\in S_t\}\) may update only \(s_{m,t+1}\), after every output
tree in round \(t\) succeeds. Consequently, a current gain \(G_{j,m,t}\) cannot affect its
own same-round FGT scale \(s_{m,t}\).
\end{proof}

\section{Empirical-Bernstein Boundary and Fixed Scale}
\label{app:eb-boundary}

\paragraph{Fixed scale.}
After native preprocessing fixes \(p\), fix \(n_{\mathrm{ref}}>0\) (default 300) and
choose \(z>0\) and \(\rho>0\) by
\begin{equation}
z-\log(1+z)=2\log(p/\delta),
\qquad \rho=\frac{n_{\mathrm{ref}}}{4z}.
\label{eq:mixture-scale}
\end{equation}
The left side increases strictly from zero to infinity, so \(z\) exists
uniquely. This
\emph{normal-reference scale} \(\rho\) uses the normal-mixture calibration convention
of \citet[Proposition~3]{howard2021confidence} at the reference variance
\(n_{\mathrm{ref}}/4\). It is a fixed design choice, not an optimum for the
gamma-exponential boundary \(u_\rho(v;\alpha)\). The value \(n_{\mathrm{ref}}/4\) is not an upper bound on
the observed squared-error sum \(V_{j,t}\). Coverage holds for every fixed positive
\(\rho\). The reference \(n_{\mathrm{ref}}\) may be configured before training, but is never
chosen from observed gains, losses, or stopping times. The stopping constants
\(\epsilon=0.001\), \(P=10\), and their derived \(n_\star=10000\) are separate
from this confidence calibration.

\paragraph{Boundary.}
For \(0\leq\lambda<1\), write
\(\psi_E(\lambda)=-\log(1-\lambda)-\lambda\). Let
\(P(a,x)=\gamma(a,x)/\Gamma(a)\) be the regularized lower incomplete gamma
function. Mix \(\exp\{\lambda s-\psi_E(\lambda)v\}\) against the density
\[
f_\rho(\lambda)=
\frac{\rho^\rho}{\Gamma(\rho)P(\rho,\rho)}
(1-\lambda)^{\rho-1}e^{-\rho(1-\lambda)},\qquad 0<\lambda<1.
\]
This is the distribution of \(1-G\) conditional on \(G<1\), for a gamma
variable \(G\) with shape and rate \(\rho\). For every real \(s\) and
\(v\geq0\), define
\[
M_\rho(s,v)=\int_0^1 e^{\lambda s-\psi_E(\lambda)v}
f_\rho(\lambda)\,d\lambda.
\]
For \(s\geq0\), the domain needed for boundary inversion, integration gives
\begin{equation}
M_\rho(s,v)=
\frac{\rho^\rho\Gamma(v+\rho)P(v+\rho,s+v+\rho)e^{s+v}}
{\Gamma(\rho)P(\rho,\rho)(s+v+\rho)^{v+\rho}}.
\label{eq:gamma-mixture}
\end{equation}
For \(v\geq0\) and \(\alpha\in(0,1)\), define \(u_\rho(v;\alpha)\)
as the unique nonnegative solution of
\(M_\rho(u_\rho(v;\alpha),v)=1/\alpha\). Uniqueness follows because the
mixture \(M_\rho(s,v)\) is continuous and strictly increasing in \(s\), starts at a value of at most one
at \(s=0\), and diverges as \(s\) increases. The boundary \(u_\rho(v;\alpha)\) is nondecreasing
in \(v\), since every mixture integrand decreases with \(v\).

\paragraph{Numerical evaluation.}
The implementation evaluates the log mixture \(\log M_\rho(s,v)\) and brackets its root, returning
the upper endpoint with an outward numerical margin. It caches boundaries \(u_\rho(v^+;\alpha)\)
on an upward-rounded logarithmic grid for \(v\): writing
\(e=\lfloor\log_2(1+v)\rfloor\), use
\(\Delta=2^e/128\) and
\(v^+=\lceil(1+v)/\Delta\rceil\Delta-1\geq v\).
Monotonicity makes \(u_\rho(v^+;\alpha)\) conservative for the exact boundary \(u_\rho(v;\alpha)\).
Division still uses the current \(N\), including when a new observation adds
zero squared error. Unsupported numerical inputs return the full interval \([0,1]\)
rather than a narrowed bound. Numerical tests compare the implementation
against high-precision roots. These tests are not a formal certification of
floating-point special functions.
All coverage and selection guarantees also hold with any conservative
numerical radius \(\widetilde r\geq r\) in place of \(r\), provided the
endpoints \(L_{j,t},U_{j,t}\), interval point estimates \(c_{j,t}\), half-widths \(h_{j,t}\), and selected sets \(S_t,S_t^{(f)}\) are all
computed from \(\widetilde r\). The selection-gap bounds then use these
implemented radii \(\widetilde r\) and half-widths \(h_{j,t}\).

\section{Proof of Theorem~\ref{thm:coverage}}
\label{app:coverage}

\begin{proof}
Condition throughout on \(\mathcal F_1\). Under the resulting conditional law,
\(p\), \(\alpha=\delta/(2p)\), and \(\rho>0\) are fixed.

\paragraph{Predictable squared-error control.}
Fix feature \(j\), suppress its index, and put
\(\widehat Y_t=I_t a_t\), \(D_t=Y_t-\nu_t\), and
\(m_t=\widehat Y_t-\nu_t\). Both \(\widehat Y_t\) and \(m_t\) are
predictable, \(\E[D_t\mid\mathcal F_{t-1}]=0\), and
\(D_t-m_t=Y_t-\widehat Y_t\in[-1,1]\).
The empirical-Bernstein exponential inequality used in
\citet[Theorem~4 and Appendix~A.8]{howard2021confidence} gives, for
\(0\leq\lambda<1\) and \(x\geq-1\),
\[
e^{\lambda x-\psi_E(\lambda)x^2}\leq1+\lambda x.
\]
Consequently,
\[
\E\!\left[e^{\lambda D_t-\psi_E(\lambda)(Y_t-\widehat Y_t)^2}
\mid\mathcal F_{t-1}\right]
\leq e^{\lambda m_t}(1-\lambda m_t)\leq1.
\]
Since \(V_u=\sum_{t=2}^u(Y_t-\widehat Y_t)^2\), multiplication over
rounds makes \(\exp\{\lambda S_u-\psi_E(\lambda)V_u\}\) a nonnegative
supermartingale starting at one, where
\(S_u=\sum_{t=2}^u D_t\). Applying the same argument to
\(-Y_t\) and \(-\widehat Y_t\) gives the corresponding supermartingale
for \(-S_u\), with exactly the same \(V_u\).

\paragraph{Mixture, inversion, and joint coverage.}
Integration against the fixed density \(f_\rho\) preserves the
supermartingale property by conditional Tonelli's theorem. Ville's inequality
therefore bounds the probability that either
\(M_\rho(S_u,V_u)\) or \(M_\rho(-S_u,V_u)\) ever reaches \(1/\alpha\)
by at most \(2\alpha\). Monotonicity in the first argument yields
\(\lvert S_u\rvert<u_\rho(V_u;\alpha)\) simultaneously at all rounds
outside these events. For \(N_u>0\), the identity
\[
S_u=A_u-\sum_{t=2}^u\nu_t
=N_u(\widehat\mu_u-\overline\nu_u)
\]
gives the radius \(r_u\) in Equation~\eqref{eq:boundary}. A conditional union bound
over the fixed \(p\) features has total error at most
\(2p\alpha=\delta\). Random observation counts \(N_u\) and adaptive stopping cause
no additional error allocation because the event already covers every round.
No independence across features, rounds, native EFB groups, GOSS draws, or
multiclass outputs is used.
\end{proof}

\section{Proofs for Feature Selection}
\label{app:mcb}

\paragraph{Relation to feature-rank inference.}
Statistical racing uses uncertainty to allocate observations adaptively
\citep{maron1994hoeffding,busafekete2009ucb}.
MABSplit allocates samples to candidate splits within a tree node
\citep{tiwari2022mabsplit}, whereas \craft{} accumulates root-gain evidence across
boosting rounds.
Simultaneous rank intervals yield confidence sets containing top-ranked features
\citep{neuhof2024ranking}, while sequential sampling procedures can verify
feature-importance rankings \citep{goldwasser2025rankings}. These guarantees
concern feature-importance targets under their respective sampling assumptions.
Here, joint coverage of accumulated expected gains \(\overline\nu_{j,t-1}\)
implies possible-best retention for the Endogenous rule and an approximate
top-\(K\) gap bound for the Exogenous rule. The latter does not assert exact
recovery of the top-\(K\) set.

\begin{proof}[Proof of Proposition~\ref{prop:sharp}]
Write \(C_t=\max_cL_{c,t}\).

Suppose first that \(U_{j,t}<C_t\). Choose \(c\) with
\(L_{c,t}=C_t\). Every \(v\in\mathcal C_t\) satisfies
\[
v_j\leq U_{j,t}<L_{c,t}\leq v_c,
\]
so feature \(j\) cannot maximize any parameter vector \(v\) in the rectangular
confidence region \(\mathcal C_t\).

Conversely, suppose \(U_{j,t}\geq C_t\). Define \(v\in\mathbb R^p\) by
\[
v_j=U_{j,t},
\qquad
v_c=L_{c,t}\quad(c\neq j).
\]
The intervals \([L_{j,t},U_{j,t}]\) are nonempty, so \(v_j\in[L_{j,t},U_{j,t}]\), and all other
coordinates also lie in their intervals \([L_{c,t},U_{c,t}]\). Moreover,
\(v_c=L_{c,t}\leq C_t\leq U_{j,t}=v_j\) for every \(c\neq j\). Thus
\(v\in\mathcal C_t\) and feature \(j\) is a maximizer of \(v\). The two sets
are therefore equal. Since every member of \(S_t\) is a maximizer for at least
one allowed vector \(v\), deleting any member would remove a possible-best feature.
This proves the stated sharpness relative to \(\mathcal C_t\).
\end{proof}

\begin{proof}[Proof of Theorem~\ref{thm:mcb}]
Let \(\Omega_\delta\) be the joint time-uniform coverage event \(\Omega_\delta\) from
Theorem~\ref{thm:coverage}. This event has conditional probability at least
\(1-\delta\). Fix any boosting round \(t\geq21\) and work
on \(\Omega_\delta\). Round 2 selected every feature, so
\(N_{j,t-1}>0\) for all \(j\).

\paragraph{A useful bounded-interval sandwich.}
Coverage gives
\(|\widehat\mu_{j,t-1}-\overline\nu_{j,t-1}|\leq r_{j,t-1}\), and
\(\overline\nu_{j,t-1}\in[0,1]\). From the definition of the bounded
endpoints \(L_{j,t},U_{j,t}\),
\begin{equation}
L_{j,t}\leq\overline\nu_{j,t-1}\leq U_{j,t},
\qquad
L_{j,t}\geq\overline\nu_{j,t-1}-2r_{j,t-1},
\qquad
U_{j,t}\leq\overline\nu_{j,t-1}+2r_{j,t-1}.
\label{eq:bounded-sandwich}
\end{equation}
Indeed,
\(\widehat\mu_{j,t-1}-r_{j,t-1}\leq\overline\nu_{j,t-1}\) and
\(0\leq\overline\nu_{j,t-1}\), which proves the first lower-endpoint
inequality. The other coverage inequality and
\(\overline\nu_{j,t-1}\leq1\) give the upper-endpoint inequality. Also,
\(\widehat\mu_{j,t-1}\geq\overline\nu_{j,t-1}-r_{j,t-1}\) implies
\(L_{j,t}\geq\overline\nu_{j,t-1}-2r_{j,t-1}\), while
\(\widehat\mu_{j,t-1}\leq\overline\nu_{j,t-1}+r_{j,t-1}\) implies
\(U_{j,t}\leq\overline\nu_{j,t-1}+2r_{j,t-1}\).

\paragraph{1. No true maximizer is omitted.}
Let \(j^\star\in\mathcal M_t^\star\). By
Equation~\eqref{eq:bounded-sandwich},
\(U_{j^\star,t}\geq\overline\nu_{j^\star,t-1}=\nu_t^\star\). For every
feature \(c\),
\(L_{c,t}\leq\overline\nu_{c,t-1}\leq\nu_t^\star\), so
\(C_t=\max_cL_{c,t}\leq\nu_t^\star\). Hence
\(U_{j^\star,t}\geq C_t\), and \(j^\star\in S_t\). Since this holds for
every true maximizer \(j^\star\in\mathcal M_t^\star\),
\[
\mathcal M_t^\star\subseteq S_t,
\qquad
K_t\geq|\mathcal M_t^\star|.
\]

\paragraph{2. Every omitted feature is strictly suboptimal.}
If \(j\notin S_t\), then \(U_{j,t}<C_t\). Choose \(c\) attaining
\(C_t=L_{c,t}\). Since \(U_{c,t}\geq L_{c,t}=C_t\), feature \(c\) is
retained. Equation~\eqref{eq:bounded-sandwich} then gives
\[
\overline\nu_{j,t-1}
\leq U_{j,t}
<C_t=L_{c,t}
\leq\overline\nu_{c,t-1},
\]
which is the claimed strict comparison.

\paragraph{3. Approximate-best guarantee for each retained feature.}
Choose \(j^\star\in\mathcal M_t^\star\) such that
\(r_{j^\star,t-1}=r_t^\star\). If \(j\in S_t\), then
\[
U_{j,t}\geq C_t\geq L_{j^\star,t}
\geq\nu_t^\star-2r_t^\star,
\]
where the final inequality is
Equation~\eqref{eq:bounded-sandwich}. The same equation also gives
\(U_{j,t}\leq\overline\nu_{j,t-1}+2r_{j,t-1}\). Combining the two displays
yields
\[
0\leq\nu_t^\star-\overline\nu_{j,t-1}
\leq2r_{j,t-1}+2r_t^\star,
\]
which is Equation~\eqref{eq:retained-bound}. Summing over \(j\in S_t\) and
dividing by \(K_t\) gives
\[
0\leq\nu_t^\star-
\frac1{K_t}\sum_{j\in S_t}\overline\nu_{j,t-1}
\leq2r_t^\star+
\frac2{K_t}\sum_{j\in S_t}r_{j,t-1}.
\]
The minimum-radius maximizer \(j^\star\) is retained by part 1, so
\(r_t^\star\leq\max_{j\in S_t}r_{j,t-1}\), and the selected-set average of the
radii \(r_{j,t-1}\) is bounded by the same maximum. Therefore the right-hand side is at most
\(4\max_{j\in S_t}r_{j,t-1}\), proving
Equation~\eqref{eq:mcb-average}.

\paragraph{4. Exact recovery.}
If \(K_t=1\), part 1 places the nonempty maximizer set
\(\mathcal M_t^\star\) inside a singleton, so the sole retained feature is the
unique maximizer. If \(\mathcal M_t^\star=\{1,\ldots,p\}\), part 1 directly
gives \(S_t=\mathcal M_t^\star\).

Now suppose \(\mathcal M_t^\star\neq\{1,\ldots,p\}\) and the separation
condition in Equation~\eqref{eq:exact-mcb} holds. Let
\[
L_t^\star=\max_{c\in\mathcal M_t^\star}L_{c,t}.
\]
For every nonmaximizer \(j\notin\mathcal M_t^\star\),
\(L_{j,t}\leq U_{j,t}<L_t^\star\). Hence the global lower-endpoint maximum is
\(C_t=L_t^\star\), and every nonmaximizer satisfies \(U_{j,t}<C_t\) and is
omitted. Every maximizer \(j\in\mathcal M_t^\star\) is retained by part 1, so
\(S_t=\mathcal M_t^\star\).

All four conclusions hold for every boosting round \(t\geq21\) on
the single event \(\Omega_\delta\), which completes the proof.
\end{proof}

\begin{proof}[Proof of Corollary~\ref{cor:adaptive-top}]
Fix a round covered by Theorem~\ref{thm:mcb} and work on the same joint
coverage event \(\Omega_\delta\). By optimality of \(S_t^{\star,K_t}\),
\[
\sum_{j\in S_t^{\star,K_t}}\overline\nu_{j,t-1}
\geq
\sum_{j\in S_t}\overline\nu_{j,t-1},
\]
which proves the lower bound in Equation~\eqref{eq:adaptive-top}.

The selection rule retains every feature with \(U_{j,t}\geq C_t\) and omits
only features with \(U_{j,t}<C_t\). Therefore every selected upper endpoint \(U_{j,t}\), \(j\in S_t\), is
strictly larger than every omitted upper endpoint \(U_{j,t}\), \(j\notin S_t\). All ties at the boundary \(C_t\) are
retained. Let
\[
A=S_t^{\star,K_t}\setminus S_t,
\qquad
B=S_t\setminus S_t^{\star,K_t}.
\]
The two sets \(A,B\) have the same size. If they are empty, the result is
immediate. Otherwise choose any bijection \(\pi:A\to B\). For each
\(j\in A\), feature \(j\) is omitted and \(\pi(j)\) is selected, so coverage
and Equation~\eqref{eq:bounded-sandwich} imply
\[
\overline\nu_{j,t-1}
\leq U_{j,t}
<U_{\pi(j),t}
\leq\overline\nu_{\pi(j),t-1}+2r_{\pi(j),t-1}.
\]
Since the common elements of \(S_t^{\star,K_t}\) and \(S_t\) cancel,
\begin{align*}
&\sum_{j\in S_t^{\star,K_t}}\overline\nu_{j,t-1}
-\sum_{j\in S_t}\overline\nu_{j,t-1}\\
&\qquad=
\sum_{j\in A}
\left(\overline\nu_{j,t-1}
-\overline\nu_{\pi(j),t-1}\right)\\
&\qquad\leq
2\sum_{i\in B}r_{i,t-1}
\leq2K_t\max_{i\in S_t}r_{i,t-1}.
\end{align*}
Dividing by \(K_t\) proves the upper bound in
Equation~\eqref{eq:adaptive-top}. The set \(S_t\) and its selection size \(K_t\) are
measurable with respect to the committed past \(\mathcal F_{t-1}\), and this exchange argument is
deterministic on the already joint event \(\Omega_\delta\). No further union bound over
realized values of \(K_t\) is required.
\end{proof}

\begin{proof}[Proof of Theorem~\ref{thm:exogenous-k}]
Work on the joint event \(\Omega_\delta\) from Theorem~\ref{thm:coverage}. Fix \(t\geq3\), and
write
\[
A=S_t^{\star,K_f}\setminus S_t^{(f)},
\qquad
B=S_t^{(f)}\setminus S_t^{\star,K_f}.
\]
These sets \(A,B\) have equal size. Choose any bijection \(\pi:A\to B\).
Because \(S_t^{(f)}\) contains the \(K_f\) largest interval point estimates \(c_{j,t}\),
\(c_{j,t}\leq c_{\pi(j),t}\) for every \(j\in A\). Coverage and the known
target range \([0,1]\) imply
\(\overline\nu_{j,t-1}\in[c_{j,t}-h_{j,t},c_{j,t}+h_{j,t}]\), so
\[
\overline\nu_{j,t-1}-\overline\nu_{\pi(j),t-1}
\leq c_{j,t}+h_{j,t}-c_{\pi(j),t}+h_{\pi(j),t}
\leq h_{j,t}+h_{\pi(j),t}.
\]
Canceling common members and summing proves the middle bound in
Equation~\eqref{eq:exogenous-k-bound}. Since \(|A|=|B|\leq K_f\) and
\(0\leq h_{j,t}\leq1/2\), the remaining upper bounds follow. The lower
bound follows from the optimality of \(S_t^{\star,K_f}\). The argument is
deterministic on the joint coverage event \(\Omega_\delta\) and applies at every selection
round, so no further union bound over rounds or selected sets \(S_t^{(f)}\) is needed.
\end{proof}

\section{Training-Loss Stopping and Returned Endpoint}
\label{sec:stopping}

The RQ1 endogenous controller uses only the complete training set. There is
no stopping-validation set, and final-test data are unavailable to it.
Let \(F_0\) be the objective's training-derived initial predictor \(F_0\) before any
boosting tree, and let \(F_t\) be the predictor \(F_t\) after \(t\) successfully
completed boosting rounds. Write \(L_t\) for its prespecified empirical
training loss \(L_t\): native binary or multiclass log loss, or RMSE for regression.
The objective transform, weights, and aggregation remain fixed.

Evaluate \(L_0\) once. The protocol rejects \(L_0=0\) as a degenerate case
rather than returning a zero-tree model. For finite \(L_0>0\), set
\[
\ell_t=\frac{L_t}{L_0},\qquad \epsilon=0.001,\qquad P=10.
\]
Thus \(\ell_0=1\), and the fixed raw-scale improvement threshold \(\eta\) is
\(\eta=\epsilon L_0\), not a fraction of a moving incumbent. Nonfinite or
negative losses \(L_t\), or a nonpositive or nonfinite computed threshold \(\eta\), are
failures rather than plateau evidence.

The stopping controller starts after round 2, even during the endogenous
full-feature initialization. Set
\[
a_2=\min\{\ell_0,\ell_1,\ell_2\},\qquad h_2=2.
\]
The anchor \(a_t\) controls clock resets, not checkpoint selection.
For every completed \(t\geq3\), first update
\begin{equation}
(a_t,h_t)=
\begin{cases}
(\ell_t,t),&\ell_t<a_{t-1}-\epsilon,\\
(a_{t-1},h_{t-1}),&\text{otherwise},
\end{cases}
\label{eq:stopping-update}
\end{equation}
and only then test \(t-h_t\geq P\). Equality at the threshold \(a_{t-1}-\epsilon\) does not reset
the clock. The stopping round \(\tau\) is
\begin{equation}
\tau=\inf\{t\geq3:t-h_t\geq P\}.
\label{eq:stopping-time}
\end{equation}
The measured endogenous fit returns \(F_\tau\), not the loss minimizer over
earlier rounds. In RQ1, its untimed pre-run reports the round count \(t^{(0)}\) and average
selected feature fraction \(\bar f\) but is discarded without test prediction.
No external round cap is supplied to an endogenous fit. Fixed-horizon
comparators instead return the model at their prescribed round count.
Passive patience bookkeeping does not shorten those fits.

\begin{theorem}[Finite training-loss stopping]
\label{thm:stopping}
Assume \(L_0\) is finite and positive and native training can continue while
producing a finite nonnegative training-loss sequence \((L_t)\). For
\(\epsilon\in(0,1)\), integer \(P\geq1\), and
Equations~\eqref{eq:stopping-update}--\eqref{eq:stopping-time}:
\begin{enumerate}
\item \(\tau\) is a stopping time for the completed-round training history \(\mathcal F_t\).
\item
\[
\tau\leq2+P\left\lceil\frac1\epsilon\right\rceil;
\]
\item for every completed \(t\geq2\),
\[
0\leq a_t-\min_{0\leq s\leq t}\ell_s\leq\epsilon;
\]
\item at a patience stop, \(\ell_s\geq a_\tau-\epsilon\) for
\(s=\tau-P+1,\ldots,\tau\).
\end{enumerate}
For the fixed \(\epsilon=0.001\) and \(P=10\), the deterministic upper bound is
10,002 completed rounds.
\end{theorem}

\begin{proof}
Every update and decision uses only the completed-round history \(\mathcal F_t\), proving the
stopping-time statement. Initially \(0\leq a_2\leq1\), and every reset decreases
the anchor \(a_t\) by strictly more than \(\epsilon\), so if \(J\) resets occur then
\(J<1/\epsilon\), hence \(J+1\leq\lceil1/\epsilon\rceil\). There are at most
\(P\) completed rounds before the first reset, between successive resets, and
between the final reset and stopping because a qualifying improvement on the
\(P\)-th round is processed first. This gives the finite bound. Induction on
Equation~\eqref{eq:stopping-update} gives the anchor inequality. During the
final \(P\) rounds no reset occurs and the anchor \(a_t\) is constant, which gives the
last assertion.
\end{proof}

The bound is neither a recommended training length nor a certificate that
later unevaluated rounds cannot improve. The anchor inequality concerns the
internal anchor \(a_t\), not the returned model's loss \(L_\tau\). The
\(n_\star=P\lceil1/\epsilon\rceil=10000\) term describes the stopping rule,
not the separate confidence reference \(n_{\mathrm{ref}}=300\). Training
is not instructed to reach either value. The implementation checks the
10,002-round limit as an integrity condition after testing patience, not as
an independently chosen training budget.

Because Theorem~\ref{thm:coverage} is joint over all completed rounds,
its feature statements remain valid through \(\tau\) without another error
allocation. If native training cannot complete another round before the
required endpoint, the fit is unavailable rather than a successful patience
stop. A completed round that reaches patience or a prescribed fixed endpoint
is accepted. Cancellation, timeout, and numerical failure remain separate
from normal stopping. Predictive quality is evaluated on the held-out test
set after the endpoint models are frozen.

\section{Algorithm and Execution Details}
\label{app:round-order}

At the beginning of training, after native Dataset construction has fixed the
feature map and revealed the feature count \(p\),
\craft{} computes the per-tail error allocation \(\alpha\) and the positive
mixture scale \(\rho\) defined in
Appendix~\ref{app:eb-boundary}. Both are frozen before round 1. The calibration
uses the prespecified reference count \(n_{\mathrm{ref}}\), not the stopping tolerance \(\epsilon\), patience \(P\),
requested or realized training length, observed gains, or losses.
The fixed feature map, native EFB bundling map, and storage representation are
included in \(\mathcal F_1\) and are not recalibrated during training.
RQ1 endogenous fits receive no external round limit, while fixed-horizon fits
receive the endpoint prescribed by their experimental protocol. Neither
endpoint enters the mixture scale \(\rho\).

\begin{algorithm}[htbp]
\caption{CRAFT-GBM}
\label{alg:craft}
\textbf{Inputs:} fixed features \([p]=\{1,\ldots,p\}\) and canonical identifiers,
error level \(\delta\in(0,1)\), fraction \(f\in(0,1]\), native learner,
and endpoint controller.\\
\textbf{Output:} model at the controller's accepted endpoint.
\begin{enumerate}
\item \textbf{Initialize:}
\(A_j=N_j=V_j=0\) for \(j\in[p]\), \(s_m=\bot\) for \(m\in[M]\), and
\(\alpha=\delta/(2p)\). Freeze \(\rho\) by Equation~\eqref{eq:mixture-scale}.
Set \(K_f=\max\{1,\lfloor fp+1/2\rfloor\}\) and \(t=1\).
\item \textbf{While} the controller requests round \(t\):
\begin{enumerate}
\item \textbf{Select from committed history.} For \(t\geq3\), compute
\((L_j,U_j)\) by Equations~\eqref{eq:boundary} and~\eqref{eq:bounded-interval},
and set the interval point estimate \(c_j=(L_j+U_j)/2\).
Set
\[
S=\begin{cases}
[p],&t\leq2\ \text{or}\ (f=1,\ t\leq20),\\
\{j:U_j\geq\max_c L_c\},&t\geq21,\ f=1,\\
\operatorname{First}_{K_f}\!\bigl(\operatorname{sort}_{j\in[p]}(-c_j,N_j,\mathrm{id}(j))\bigr),
&t\geq3,\ f<1.
\end{cases}
\]
Here \(\mathrm{id}(j)\) is the canonical feature identifier \(\mathrm{id}(j)\). Sorting is
ascending lexicographic, and \(\operatorname{First}_{K_f}\) returns the first
\(K_f\) feature indices.
\item \textbf{Build a tentative round.} Fix \(S\) before GOSS and use it for all
\(M\) output trees. Record \(G_{j,m,t}\) from Equation~\eqref{eq:feature-gain}
for every \(j\in S\) and \(m\in[M]\).
\item \textbf{If the native round is discarded or an exception occurs:}
abort pending selector updates and restore its committed mask, leaving
statistics \(A_j,N_j,V_j\), scales \(s_m\), and the endpoint controller unchanged. Do not advance
\(t\). Native code handles model termination, and exceptions fail the fit
rather than resuming from a partially built round.
\item \textbf{Otherwise, stage the updates.} For \(t\geq2\) and \(j\in S\),
using the pre-round scales \(s_m\) and pre-update statistics \(A_j,N_j,V_j\), compute
\[
R_j=\frac1M\sum_{m=1}^M\phi(G_{j,m,t};s_m),\qquad
a_j=\frac{1/2+A_j}{1+N_j},
\]
\[
(A_j^+,N_j^+,V_j^+)
=(A_j+R_j,\ N_j+1,\ V_j+(R_j-a_j)^2).
\]
For \(t=1\) or \(j\notin S\), set
\((A_j^+,N_j^+,V_j^+)=(A_j,N_j,V_j)\).
For each output \(m\), let \(H_m\) be the multiset of positive
\(G_{j,m,t}\) values, one per selected feature \(j\in S\), and set
\[
s_m^+=\begin{cases}
\operatorname{median}(H_m),&H_m\ne\varnothing,\\
s_m,&H_m=\varnothing.
\end{cases}
\]
\item \textbf{Commit} the completed model \(F_t\), \(S_t=S\), \(K_t=|S|\), and
\((A,N,V,s)\leftarrow(A^+,N^+,V^+,s^+)\).
Update the endpoint controller from this round and set \(t\leftarrow t+1\).
\end{enumerate}
\end{enumerate}
\end{algorithm}

Round 1 initializes scales \(s_m\) only, and round 2 supplies the first confidence
observation \(R_{j,2}\) for every feature. Endogenous fits remain full-feature through
round 20 if reached, while updating statistics \(A_j,N_j,V_j\) normally. A selected feature with zero gain still
increments its observation count \(N_j\), whereas an omitted feature changes none
of \(A,N,V\). The median \(\operatorname{median}(H_m)\) uses the midpoint for an even number of values.
The training-loss controller in Appendix~\ref{sec:stopping} accepts the
patience-triggering round \(\tau\) and returns that model \(F_\tau\). Fixed-horizon fits return
their prescribed endpoint. An incomplete attempt is not a successful stop.
\FloatBarrier

\section{Native Execution}
\label{app:native-execution}

Native LightGBM constructs EFB groups before boosting
\citep{ke2017lightgbm}. Loading a binary Dataset restores the same fixed
representation. Let \(\mathcal E=\{1,\ldots,B\}\) index these groups and let
\(D_b\subseteq\{1,\ldots,p\}\) contain the valid features stored in
group \(b\). For any selected feature set \(S\), define its direct
EFB-group closure \(\mathcal E(S)\)
\begin{equation}
\mathcal E(S)=\{b\in\mathcal E:D_b\cap S\neq\varnothing\}.
\label{eq:efb-closure}
\end{equation}
In the validated forced-column-wise CPU path, a group is invoked exactly when
it belongs to this closure \(\mathcal E(S)\). The closure \(\mathcal E(S)\) is nevertheless a coarse work
descriptor: an ordinary packed group constructs its shared histogram when any
member is enabled, and the unchanged multi-value subcolumn heuristic may
process all member columns in an implicated group. This physical processing
does not make features outside the exact mask \(S\) eligible or update their
statistics. \craft{} neither rebuilds
EFB nor changes native binning, packed storage, subset construction, or
histogram arithmetic.

Native GOSS selects and reweights rows \citep{ke2017lightgbm}. The selected
feature mask \(S\) and its direct group closure \(\mathcal E(S)\) are fixed before the current row
draw. Rows and weights then determine root gains \(G_{j,m,t}\). No per-feature row sample or
additional histogram pass is introduced.

\section{Model Parameters of CRAFT-GBM}
\label{sec:parameters}

\begin{table}[ht]
\caption{Model parameters and conventions of CRAFT-GBM}
\label{tab:parameters}
\begin{center}
\begin{tabular}{@{}>{\raggedright\arraybackslash}p{1.15in}>{\raggedright\arraybackslash}p{1.15in}>{\raggedright\arraybackslash}p{2.65in}@{}}
\toprule
Quantity & Value & Role and limitation\\
\midrule
Enable selector & \texttt{true} for \craft{} &
Set \codefield{adaptive_feature_selection=true}. The native default is \texttt{false}.\\
Feature count & \(p\) &
Number of features retained by native preprocessing, fixed before covered gains.\\
Joint-confidence error level & \(\delta=0.05\) (default) &
Joint-confidence error level \(\delta\) over the fixed features for all boosting
rounds, fixed at the beginning of training.\\
Per-tail error level & \(\alpha=\delta/(2p)\) &
Derived from the fixed feature count \(p\) and joint-confidence error level \(\delta\), not
from observed gains.\\
Boosting-round endpoint & endogenous or protocol-fixed &
RQ1 endogenous fits use training-loss patience \(P\) without an external cap.
Fixed-horizon fits use their prescribed endpoint. Neither determines
the feature-confidence boundary \(u_\rho(v;\alpha)\).\\
\codefield{feature_fraction} & \(1.0\) (default) or \(f\in(0,1)\) &
With the selector enabled, \(1.0\) invokes the endogenous rule for the selection size \(K_t\), while \(f<1\) invokes
Exogenous \(K_f\) as defined in Equation~\eqref{eq:exogenous-k}.\\
Confidence reference & \(n_{\mathrm{ref}}=300\) (default) &
A scale reference \(n_{\mathrm{ref}}\) configurable before training, not a training horizon.\\
Mixture scale & \(\rho=n_{\mathrm{ref}}/(4z)\) &
Derived by Equation~\eqref{eq:mixture-scale} and frozen before round 1.
The training horizon, gains, and losses are excluded.\\
Stopping rule & \(\epsilon=0.001,\ P=10,\ n_\star=10000\) &
Fixed training-loss improvement tolerance \(\epsilon\) and patience \(P\). The derived
\(n_\star=P\lceil1/\epsilon\rceil=10000\) is their deterministic
stopping-bound term, not a coverage horizon or execution target.\\
\bottomrule
\end{tabular}
\end{center}
\end{table}

Table~\ref{tab:parameters} lists the model parameters and conventions. The experiments fix
\(n_{\mathrm{ref}}=300\) in RQ1 and 1000 in RQ2.

\paragraph{Supported execution.}
The extension supports fresh, continuous, single-machine fits with native
objectives and \codefield{boosting=gbdt}, \codefield{device_type=cpu},
\codefield{tree_learner=serial}, \codefield{force_col_wise=true},
\codefield{force_row_wise=false}, \codefield{enable_bundle=true},
\codefield{is_enable_sparse=true}, and \codefield{feature_fraction_bynode=1.0}.
Row sampling is native GOSS or no sampling, with ordinary and query bagging
disabled. Unsupported configurations include custom objectives, GPU or distributed
training, warm starts, native early stopping, linear trees, quantized gradients,
extra trees, CEGB, monotone or interaction constraints, feature contributions,
and forced splits.

The native extension implements feature selection and confidence updates.
The supplied benchmark driver separately implements the RQ1 training-loss
stopping rule: it updates the booster one round at a time, evaluates training
loss, applies patience, and saves the completed endpoint.
Enabling the selector alone does not install this stopping rule.

The joint-confidence error level \(\delta\) may be any \(\mathcal F_1\)-measurable value fixed at
the beginning of training, but changing it after training begins to obtain a
preferred selection size \(K_t\) invalidates the stated guarantee. Each interval width \(2h_{j,t}\)
uses the fixed feature count \(p\) and that feature's actual selection history.
The feature-selector controls are \(\delta\)
(\codefield{adaptive_feature_confidence_delta}), native
\texttt{feature\_fraction} \(f\), and the confidence
reference \(n_{\mathrm{ref}}\) \codefield{adaptive_feature_calibration_reference_count}.
The feature count \(p\) follows from
native construction. The fraction \(f\) is fixed for the fit.
Training-time reconfiguration is not a supported way to choose a preferred
selection size \(K_t\). The internal \(\epsilon,P\) values and the fixed
mixture-scale calibration convention are disclosed
design choices rather than parameter-free scientific consequences.

\clearpage

\section{Synthetic Data Generation}
\label{app:synthetic}

The four synthetic families complement the public datasets with controlled
signal and input structures. Every generated replicate has 2,000 numerical
features and 100,000 training, 20,000 validation, and 100,000 test observations.
The release contains ten replicates per family, indexed 00--09, while the publication
benchmark retains the prespecified IDs 00--04, chosen by index before model
results were inspected. All models share the same observed arrays within a
replicate and use model seed 314159. Replicates are separate learning problems
and are never concatenated or replaced according to their outcomes. The
generated validation splits are unused, and RQ2 instead creates its inner validation
split from the training file. All four families enter RQ1 and RQ2.

Below, coordinates refer to the canonical representation before a
replicate-specific random column permutation, shared across its three splits.
All stored inputs are float32: Sparse 1/2 use sparse CSR matrices and Dense 1/2 use dense
arrays. Binary labels are uint8, whereas Dense 1 labels and all supplied oracle targets are
float64. Native preprocessing is fitted on training data only. Sparse zeros
remain observed values, with \codefield{zero_as_missing=false}.

\paragraph{Sparse 1: independent sparse Gaussian inputs.}
For each row and coordinate, independently draw
\[
B_j\sim\operatorname{Bernoulli}(a),\qquad Z_j\sim N(0,1),\qquad
X_j=B_jZ_j/\sqrt a,\qquad a=0.05.
\]
The first 100 canonical coordinates have independent signs
\(b_j\in\{-1,+1\}\), drawn uniformly once per replicate and held fixed across
splits. With \(\sigma(z)=(1+e^{-z})^{-1}\), define
\[
f_*(X)=\frac{2}{\sqrt{100}}\sum_{j=1}^{100}b_jX_j,
\qquad p_*(X)=\sigma(f_*(X)),\qquad
Y\mid X\sim\operatorname{Bernoulli}(p_*(X)).
\]
The other 1,900 coordinates are nuisance features. The Bernoulli mask \(B_j\) is
sampled entrywise, giving density \(a\) of 0.05 in expectation rather than a fixed
nonzero count. Under the ideal real-valued law, each coordinate \(X_j\) has variance
one, \(\operatorname{Var}(f_*(X))=4\), and
\(\Pr(Y=1)=1/2\). The probability of no active signal coordinate is
\(0.95^{100}\approx0.00592\).

\paragraph{Sparse 2: exclusive categorical indicators.}
Draw 100 independent fields \(C_g\), each uniform on \(\{1,\ldots,20\}\),
and release the numerical indicators \(X_{g,c}=\mathbf1\{C_g=c\}\).
Every row has exactly 100 ones, hence density 0.05. Select five signal fields
\(\mathcal G_*\) uniformly without replacement. For each, randomly permute
ten \(+1\) and ten \(-1\) category effects \(\theta_{g,c}\), once per
replicate, and set
\[
f_*(X)=\frac{2}{\sqrt5}\sum_{g\in\mathcal G_*}\theta_{g,C_g},
\qquad p_*(X)=\sigma(f_*(X)),\qquad
Y\mid X\sim\operatorname{Bernoulli}(p_*(X)).
\]
Thus 100 indicator columns are signal-bearing and 1,900 are nuisance. The
five selected effects \(\theta_{g,C_g}\), \(g\in\mathcal G_*\), are independent Rademacher variables, giving score \(f_*(X)\)
variance four and marginal class probability \(\Pr(Y=1)\) one-half. Sparse 1/2 sample rows and
splits independently conditional on their fixed replicate metadata, with
separate label streams. Their scores \(f_*(X)\) and probabilities \(p_*(X)\) are evaluated in
float64 from the stored float32 inputs \(X\) before sampling labels \(Y\). All models
receive the same numerical one-hot representation for Sparse 2.

\paragraph{Dense 1: dense nonlinear regression.}
We implement the Friedman-1 response \citep{friedman1991mars} with 1,995
additional nuisance coordinates. Draw independent \(U_j\sim
\operatorname{Uniform}(0,1)\), store \(X_j\) as their float32 values, and
compute
\[
f_*(X)=10\sin(\pi X_1X_2)+20(X_3-0.5)^2+10X_4+5X_5,
\qquad Y=f_*(X)+\xi,
\]
where \(\xi\sim N(0,1)\) is independent of the inputs. Each split uses
separate covariate and noise streams. The response function \(f_*(X)\) is evaluated in
float64 after input rounding, so the saved \(f_*\) corresponds to the
released inputs. The signal-coordinate map preserves formula order when
undoing the column permutation.

\paragraph{Dense 2: dense classification with redundant features.}
One call to \codefield{make_classification} in scikit-learn 1.8.0
\citep{pedregosa2011scikit} generates the complete 220,000-row pool for each
replicate. The parameters are 2,000 features, five original informative
coordinates, 15 redundant linear combinations, no repeated coordinates, two
equally weighted classes, and 16 clusters per class, with
\codefield{class_sep=1}, \codefield{hypercube=True},
\codefield{shift=0}, \codefield{scale=1}, and \codefield{shuffle=False}.
The remaining 1,980 coordinates are independent standard-normal nuisance
features. The 32 clusters occupy all vertices of \(\{-1,+1\}^5\), initially
with 6,875 rows per cluster. Within each cluster, standard-normal coordinates
undergo a random linear transformation before translation to the vertex,
while the redundant coordinates use a shared random linear map. Both types of map
have entries drawn independently from \(\operatorname{Uniform}(-1,1)\).
The pinned implementation assigns class \(k\bmod2\) to vertex \(k\) in
binary order, \(k=0,\ldots,31\), giving the fifth canonical coordinate a
class-dependent mean before label randomization.
With \codefield{flip_y=0.01}, selected labels are reassigned uniformly to
either class, so the expected changed-label fraction is 0.005.

A separate outcome-blind row permutation partitions the pool into the fixed
training, validation, and test sizes, without stratification or rebalancing.
These splits are disjoint partitions of a pool with fixed initial cluster
counts, rather than independent unconditional mixture samples. Float32
storage makes the linear redundancy approximate, but all 20 informative or
redundant coordinates count as signal-bearing. Dense 2 is MADELON-style, not the
original UCI MADELON dataset, and supplies no Bayes-posterior oracle.

\paragraph{Reproducibility.}
The root seed is 20260909. NumPy streams use PCG64 initialized by
\codefield{SeedSequence} with key
\((\text{family},\text{replicate},\text{split},\text{role})\).
Family IDs are 1--4 for Sparse 1, Sparse 2, Dense 1, and Dense 2, and split IDs are 0 for metadata and
1--3 for training, validation, and test. Metadata role 0 determines the
column permutation and, for Sparse 1/2, signs \(b_j\) or field contrasts \(\theta_{g,c}\). Within splits,
Sparse 1 uses roles 0/1/2 for masks/values/labels \((B_j,Z_j,Y)\), Sparse 2 uses 0/2 for categories/labels \((C_g,Y)\),
and Dense 1 uses 0/1 for covariates/noise \((U_j,\xi)\). Dense 2 derives its scikit-learn integer seed
from the first uint32 state value of key \((4,r,0,1)\) for replicate \(r\),
and uses key \((4,r,0,2)\) for the pool-row permutation. Generation pins
Python 3.12.14, NumPy 2.3.5, SciPy 1.17.0, and scikit-learn 1.8.0 for Dense 1/2.
Sparse 1 and Dense 1 use 1,024-row generation blocks. Release records bind source and
environment information to artifact hashes and validation reports, and no
replicate is rejected for distributional fluctuations.

\paragraph{Oracle evaluation and scope.}
Oracle targets and feature-role maps are unavailable to fitting, stopping,
and feature-fraction selection. After all scheduled models for the experimental arm are frozen,
evaluation uses each replicate's \(n\) test inputs. For Sparse 1/2, let \(q_i\) be
the fitted model's predicted probability \(q_i\) at \(X_i\), clipped to
\([10^{-15},1-10^{-15}]\). We additionally report conditional cross-entropy
\[
\frac1n\sum_{i=1}^n
\bigl[-p_*(X_i)\log q_i-(1-p_*(X_i))\log(1-q_i)\bigr]
\]
and mean \(\operatorname{KL}(\operatorname{Bernoulli}(p_*(X_i))\Vert
\operatorname{Bernoulli}(q_i))\). Dense 1 reports direct known-function MSE,
\(n^{-1}\sum_i(\widehat f(X_i)-f_*(X_i))^2\), alongside test RMSE.
We do not substitute test MSE minus one for this quantity.
These diagnostics concern prediction, not confidence coverage for accumulated
transformed root gains \(\overline\nu_{j,t}\). Sparse 1/2 have additive logits \(f_*(X)\), and equal density and score
variance do not imply equal Bayes difficulty. Sparse 2's within-field exclusivity
does not establish native EFB grouping or a unique minimal indicator support.
Cross-family comparisons do not isolate a causal effect of density.

\clearpage

\section{Experimental Details and Numerical Results}
\label{app:experimental-results}

The following tables give all reported dataset--configuration combinations.
Unlike the 16-experiment main RQ1 display, these complete results retain both
+4 expansions and summarize all 18 measured experiments in RQ1. The main-paper
omission is a post-run presentation choice, not an experimental exclusion.
Both expansion sizes remain in the appendix.
The ten public datasets are HIGGS, Epsilon, Bosch, YearMSD,
APS Failure, KDD 2009, Gisette, Jannis, HAR, and CT Scan
\citep{whiteson2014higgs,libsvmEpsilon,boschFigshare,uciYearMSD,uciAPS,
kdd2009data,uciGisette,openmlJannis,uciHAR,uciCT}. Together with Sparse 1/2 and
Dense 1/2 (Appendix~\ref{app:synthetic}), they form 14 source families.
RQ1 adds two expansions each of HAR and YearMSD,
giving 18 experiments. Each has five paired measurements: model seeds for
nine fixed public splits, patient folds for CT, and generated replicates
00--04 for synthetic and expanded experiments. RQ1 reports 450 completed
training/test evaluations. The original RQ2 campaign comprises 1,470 evaluations.
RQ2 replaces the original exogenous results with 350 point-estimate-ranking
fits and 70 fresh paired Full fits. A further 770 fits supply the two ablations
and their Full references. Replaced results are not pooled with their replacements.
YearMSD, CT, Dense 1, and the YearMSD expansions are
regression experiments, while the others are classification experiments.

\begin{table}[ht]
\centering
\caption{Public datasets and fixed outer splits. C denotes classification
with log loss, and R denotes regression with RMSE. Training counts include
any inner validation split used for RQ2 tuning.}
\label{tab:public-datasets}
\begin{tabular*}{\linewidth}{@{\extracolsep{\fill}}lcrrl@{}}
\toprule
Dataset & Task & Training rows & Test rows & Outer split\\
\midrule
HIGGS & C & 10,500,000 & 500,000 & First/last rows\\
Epsilon & C & 400,000 & 100,000 & Official\\
Bosch & C & 946,997 & 236,750 & Stratified\\
YearMSD & R & 463,715 & 51,630 & First/last rows\\
APS Failure & C & 60,000 & 16,000 & Official\\
KDD 2009 & C & 40,000 & 10,000 & Stratified\\
Gisette & C & 6,000 & 1,000 & Official labeled files\\
Jannis & C & 56,102 & 27,631 & OpenML task 168792\\
HAR & C & 7,352 & 2,947 & Subject-disjoint\\
CT Scan & R & Fold-dependent & Fold-dependent & Patient-disjoint\\
\bottomrule
\end{tabular*}
\end{table}

Table~\ref{tab:public-datasets} identifies the ten public families.
Bosch uses numeric inputs only, KDD 2009 is the Appetency task, and YearMSD
denotes YearPredictionMSD. Bosch and KDD use frozen, class-stratified splits
ranked by SHA-256 with seed 271828. Gisette uses its official training file
for fitting and its labeled file named validation exclusively for final testing.
HAR is the smartphone activity dataset, with 21 training and nine test subjects.
CT Scan denotes Relative Location of CT Slices and excludes patient IDs from
the features. Its five training/test row-count pairs are
42,792/10,708, 42,784/10,716, 42,781/10,719, 42,851/10,649, and
42,792/10,708.

\paragraph{Reproducibility seeds.}
The nine fixed-split public datasets use measured model seeds 73001--73005.
CT fold \(i\in\{1,\ldots,5\}\) uses seed \(73000+i\), with one measurement
per fold. Synthetic and expanded experiments use generated replicates
00--04 and measured model seed 314159.
Every RQ1 pre-run uses model seed 271828.

The expanded experiments append four or eight independently row-permuted copies
of each source feature, giving five or nine times the original width before
native preprocessing. Their official train/test row assignments are preserved.
The permutations are label-blind but use pooled covariates, and finite-sample
split gains of the added columns need not be zero.

Absolute training times and test losses are arithmetic means with sample
standard deviations across five measurements. These are not standard errors.
CT instead uses the mean and fold range because its training folds overlap.
Loss is in its native units: natural-log loss per observation for
classification and RMSE for regression. Raw losses are not pooled across
datasets. Rounds count boosting iterations, not individual class-specific
trees. A retained fraction averages selected feature counts \(K_s\) over all rounds,
including initialization.

Separate oracle tables report Sparse 1/2 conditional cross-entropy and KL divergence,
and Dense 1 error against its known regression function \(f_*\). These use evaluation-only
generator information. They do not measure coverage of the feature-gain
confidence bounds \([L_{j,t},U_{j,t}]\).

Paired percentage changes use the geometric mean of the five candidate/reference
ratios, followed by \(100(r-1)\). They are not ratios of the arithmetic means
in the absolute-outcome tables. Brackets in paired tables apply the same
transformation to 95\% Student-\(t\) intervals on log ratios, or to CT's
descriptive fold range. Seed intervals condition on the fixed split.
Synthetic intervals vary generated problems, and expansion intervals condition
on the shared source data. RQ2 intervals also condition on its fixed tuning
split and selected settings. None incorporates uncertainty from new test samples.

Aggregate figures give equal weight to each experiment after within-experiment
geometric averaging. RQ2 resamples its 14 families, using 10,000 shared resamples
across all settings and both outcomes (PCG64 seed 20260919), with pointwise
2.5th and 97.5th percentiles. These ranges describe sensitivity to benchmark
composition conditional on the measured effects, not a joint confidence
region, a simultaneous band, or a significance test.

\paragraph{Implementation and evaluation details.}
LightGBM-based models share native preprocessing and EFB, depth 6, at most
64 leaves, maximum bin
count 255, minimum leaf size 20, L2 penalty 1, and GOSS rates 0.2/0.1
\citep{ke2017lightgbm}. XGBoost uses CPU histogram training with depth 6, and CatBoost uses
depth-6 symmetric trees. Their different samplers do not imply identical
realized feature counts. Fits run serially on fixed physical cores of an
Intel Xeon Gold 6242 server: eight cores for RQ1 and sixteen for RQ2,
identically for all models within each arm. Resource checks accept or retry
each fit separately. RQ1 fixes learning rate 0.05 and
\(n_{\mathrm{ref}}=300\), with \(\delta=0.05\). Identical lightweight 25-round timing markers are
included within each arm's training measurements, without alternative-mask
diagnostics. Untimed resource admission precedes each fit. A two-second
monitor checks residual CPU activity, I/O and memory pressure, swap activity,
and CPU affinity. Only the first accepted fit is retained. Final-test
prediction and scoring use saved models after training is complete and do
not enter the training-time comparison.

\subsection{RQ1: Endogenous and Matched-Control Results}

One untimed Endogenous CRAFT-GBM pre-run per training set supplies its round
count \(t^{(0)}\) and average feature fraction
\(y^{(0)}=(p t^{(0)})^{-1}\sum_{s=1}^{t^{(0)}}K_s\).
LightGBM-Full is the paired reference, using all features for \(t^{(0)}\)
rounds. LightGBM-Fixed uses this endpoint \(t^{(0)}\) and an integer feature count chosen
to minimize cumulative feature-count mismatch. XGBoost and CatBoost receive the
same round count \(t^{(0)}\) and fraction \(y^{(0)}\) through native sampling parameters, without
exact-count matching. The measured Endogenous CRAFT-GBM fit selects and stops
independently, without receiving the pre-run's fraction \(y^{(0)}\) or horizon \(t^{(0)}\).
Both endogenous pre-runs and measured fits use 20 full-feature rounds, if reached.
The training-loss stopping rule is unchanged.

\begingroup
\begin{longtable}{@{}lrr@{}}
\caption{RQ1: aggregate paired percentage changes. The 18 experiments receive equal weight and belong to 14 source families, and this aggregate is descriptive.}\label{tab:rq1-numerical-summary}\\
\toprule
Model & $\Delta$ time (\%) & $\Delta$ loss (\%) \\
\midrule
\endfirsthead
\multicolumn{3}{@{}l}{\tablename~\thetable{} (continued)}\\
\toprule
Model & $\Delta$ time (\%) & $\Delta$ loss (\%) \\
\midrule
\endhead
\midrule
\multicolumn{3}{r}{Continued on next page}\\
\endfoot
\bottomrule
\endlastfoot
Endogenous CRAFT-GBM & -39.74 & -1.04 \\
LightGBM-Fixed & -30.91 & -1.18 \\
XGBoost & 148.74 & 4.57 \\
CatBoost & 153.23 & 5.56 \\
\end{longtable}
\endgroup
\begingroup
\begin{longtable}{@{}lrrr@{}}
\caption{RQ1: frozen pre-run feature budgets and horizons. Values are means [minimum, maximum] across the five paired blocks per experiment. Fixed-split public datasets share one pre-run across model seeds, while other blocks use their own pre-run.}\label{tab:rq1-numerical-prerun}\\
\toprule
Dataset & $p$ & Pre-run rounds \(t^{(0)}\) & Pre-run fraction \(y^{(0)}\) (\%) \\
\midrule
\endfirsthead
\multicolumn{4}{@{}l}{\tablename~\thetable{} (continued)}\\
\toprule
Dataset & $p$ & Pre-run rounds \(t^{(0)}\) & Pre-run fraction \(y^{(0)}\) (\%) \\
\midrule
\endhead
\midrule
\multicolumn{4}{r}{Continued on next page}\\
\endfoot
\bottomrule
\endlastfoot
HIGGS & 28 & 188.0 & 59.65 \\
Epsilon & 2000 & 1128.0 & 16.40 \\
Bosch & 968 & 1339.0 & 31.64 \\
YearMSD & 90 & 270.0 & 69.77 \\
APS Failure & 170 & 364.0 & 59.90 \\
KDD 2009 & 212 & 517.0 & 22.89 \\
Gisette & 4326 & 257.0 & 39.73 \\
Jannis & 54 & 1612.0 & 74.42 \\
HAR & 561 & 111.0 & 94.21 \\
CT Scan & 378 & 343.8 [289.0, 400.0] & 50.69 [47.16, 55.50] \\
Sparse 1 & 2000 & 932.2 [910.0, 955.0] & 18.80 [18.62, 19.11] \\
Sparse 2 & 2000 & 615.6 [576.0, 656.0] & 21.42 [21.01, 21.93] \\
Dense 1 & 2000 & 171.2 [164.0, 181.0] & 43.83 [42.07, 45.12] \\
Dense 2 & 2000 & 1108.2 [967.0, 1192.0] & 19.48 [17.31, 20.97] \\
HAR +4 & 2805 & 111.2 [110.0, 112.0] & 72.26 [71.71, 72.89] \\
HAR\_N & 5049 & 111.6 [110.0, 116.0] & 71.74 [69.56, 72.68] \\
YearMSD +4 & 450 & 326.8 [299.0, 346.0] & 37.80 [36.61, 39.33] \\
YearMSD\_N & 810 & 316.8 [302.0, 326.0] & 32.99 [32.22, 33.85] \\
\end{longtable}
\endgroup
\begingroup
\begin{longtable}{@{}lrrr@{}}
\caption{RQ1: absolute outcomes for every reported configuration. Each row summarizes five replicates: mean $\pm$ sample SD, except CT, which uses mean [minimum, maximum].}\label{tab:rq1-numerical-raw}\\
\toprule
Model & Mean rounds & Training (s) & Test loss \\
\midrule
\endfirsthead
\multicolumn{4}{@{}l}{\tablename~\thetable{} (continued)}\\
\toprule
Model & Mean rounds & Training (s) & Test loss \\
\midrule
\endhead
\midrule
\multicolumn{4}{r}{Continued on next page}\\
\endfoot
\bottomrule
\endlastfoot
\addlinespace[3pt]
\multicolumn{4}{@{}l}{\textbf{HIGGS --- Binary log loss (nats/observation)}}\\*
LightGBM-Full & 188.0 & $52.91\pm 0.13$ & $0.5268\pm 1.89\!\times\!10^{-4}$ \\*
Endogenous CRAFT-GBM & 199.4 & $51.03\pm 1.49$ & $0.5288\pm 9.72\!\times\!10^{-4}$ \\*
LightGBM-Fixed & 188.0 & $49.16\pm 0.34$ & $0.5282\pm 2.00\!\times\!10^{-4}$ \\*
XGBoost & 188.0 & $75.37\pm 0.29$ & $0.5288\pm 5.08\!\times\!10^{-4}$ \\*
CatBoost & 188.0 & $78.57\pm 0.21$ & $0.5449\pm 4.22\!\times\!10^{-5}$ \\
\addlinespace[3pt]
\multicolumn{4}{@{}l}{\textbf{Epsilon --- Binary log loss (nats/observation)}}\\*
LightGBM-Full & 1128.0 & $221.55\pm 0.24$ & $0.2849\pm 1.96\!\times\!10^{-4}$ \\*
Endogenous CRAFT-GBM & 1139.6 & $54.58\pm 0.62$ & $0.2804\pm 6.29\!\times\!10^{-4}$ \\*
LightGBM-Fixed & 1128.0 & $72.21\pm 0.19$ & $0.2886\pm 1.78\!\times\!10^{-4}$ \\*
XGBoost & 1128.0 & $786.48\pm 1.42$ & $0.2904\pm 3.52\!\times\!10^{-4}$ \\*
CatBoost & 1128.0 & $198.32\pm 0.48$ & $0.3020\pm 1.05\!\times\!10^{-4}$ \\
\addlinespace[3pt]
\multicolumn{4}{@{}l}{\textbf{Bosch --- Binary log loss (nats/observation)}}\\*
LightGBM-Full & 1339.0 & $196.22\pm 0.20$ & $0.0313\pm 4.04\!\times\!10^{-5}$ \\*
Endogenous CRAFT-GBM & 1470.8 & $99.84\pm 15.18$ & $0.0314\pm 8.93\!\times\!10^{-5}$ \\*
LightGBM-Fixed & 1339.0 & $176.08\pm 0.23$ & $0.0314\pm 3.04\!\times\!10^{-5}$ \\*
XGBoost & 1339.0 & $105.97\pm 0.69$ & $0.0314\pm 2.36\!\times\!10^{-5}$ \\*
CatBoost & 1339.0 & $541.07\pm 0.99$ & $0.0317\pm 1.68\!\times\!10^{-5}$ \\
\addlinespace[3pt]
\multicolumn{4}{@{}l}{\textbf{YearMSD --- RMSE (target units)}}\\*
LightGBM-Full & 270.0 & $5.36\pm 0.02$ & $9.0259\pm 0.0022$ \\*
Endogenous CRAFT-GBM & 252.4 & $4.53\pm 0.21$ & $9.0630\pm 0.0062$ \\*
LightGBM-Fixed & 270.0 & $4.70\pm 0.02$ & $9.0191\pm 0.0037$ \\*
XGBoost & 270.0 & $8.96\pm 0.15$ & $9.0255\pm 0.0029$ \\*
CatBoost & 270.0 & $9.48\pm 0.01$ & $9.1688\pm 0.0035$ \\
\addlinespace[3pt]
\multicolumn{4}{@{}l}{\textbf{APS Failure --- Binary log loss (nats/observation)}}\\*
LightGBM-Full & 364.0 & $3.21\pm 0.02$ & $0.0210\pm 2.13\!\times\!10^{-4}$ \\*
Endogenous CRAFT-GBM & 347.2 & $2.52\pm 0.04$ & $0.0203\pm 1.02\!\times\!10^{-4}$ \\*
LightGBM-Fixed & 364.0 & $2.87\pm 0.02$ & $0.0208\pm 3.20\!\times\!10^{-4}$ \\*
XGBoost & 364.0 & $4.77\pm 0.04$ & $0.0204\pm 4.15\!\times\!10^{-4}$ \\*
CatBoost & 364.0 & $6.30\pm 0.11$ & $0.0199\pm 2.45\!\times\!10^{-4}$ \\
\addlinespace[3pt]
\multicolumn{4}{@{}l}{\textbf{KDD 2009 --- Binary log loss (nats/observation)}}\\*
LightGBM-Full & 517.0 & $3.41\pm 0.02$ & $0.1164\pm 6.01\!\times\!10^{-4}$ \\*
Endogenous CRAFT-GBM & 536.0 & $2.31\pm 0.03$ & $0.1146\pm 9.77\!\times\!10^{-4}$ \\*
LightGBM-Fixed & 517.0 & $2.47\pm 0.02$ & $0.0999\pm 6.80\!\times\!10^{-4}$ \\*
XGBoost & 517.0 & $11.64\pm 0.21$ & $0.1583\pm 4.65\!\times\!10^{-4}$ \\*
CatBoost & 517.0 & $15.37\pm 0.24$ & $0.0726\pm 3.22\!\times\!10^{-4}$ \\
\addlinespace[3pt]
\multicolumn{4}{@{}l}{\textbf{Gisette --- Binary log loss (nats/observation)}}\\*
LightGBM-Full & 257.0 & $12.51\pm 0.02$ & $0.0715\pm 0.0015$ \\*
Endogenous CRAFT-GBM & 262.2 & $7.24\pm 0.09$ & $0.0700\pm 0.0022$ \\*
LightGBM-Fixed & 257.0 & $7.56\pm 0.04$ & $0.0646\pm 6.56\!\times\!10^{-4}$ \\*
XGBoost & 257.0 & $24.06\pm 0.04$ & $0.0651\pm 0.0016$ \\*
CatBoost & 257.0 & $36.69\pm 0.33$ & $0.0809\pm 0.0018$ \\
\addlinespace[3pt]
\multicolumn{4}{@{}l}{\textbf{Jannis --- Multiclass log loss (nats/observation)}}\\*
LightGBM-Full & 1612.0 & $15.31\pm 0.05$ & $0.7677\pm 0.0016$ \\*
Endogenous CRAFT-GBM & 1559.2 & $13.34\pm 0.65$ & $0.7446\pm 0.0064$ \\*
LightGBM-Fixed & 1612.0 & $14.21\pm 0.12$ & $0.7633\pm 0.0018$ \\*
XGBoost & 1612.0 & $25.59\pm 0.10$ & $0.7042\pm 7.02\!\times\!10^{-4}$ \\*
CatBoost & 1612.0 & $25.83\pm 0.09$ & $0.6801\pm 3.98\!\times\!10^{-4}$ \\
\addlinespace[3pt]
\multicolumn{4}{@{}l}{\textbf{HAR --- Multiclass log loss (nats/observation)}}\\*
LightGBM-Full & 111.0 & $2.67\pm 0.01$ & $0.1687\pm 0.0036$ \\*
Endogenous CRAFT-GBM & 114.6 & $2.67\pm 0.04$ & $0.1694\pm 0.0017$ \\*
LightGBM-Fixed & 111.0 & $2.60\pm 0.00$ & $0.1667\pm 0.0030$ \\*
XGBoost & 111.0 & $9.62\pm 0.02$ & $0.2181\pm 0.0016$ \\*
CatBoost & 111.0 & $19.38\pm 0.06$ & $0.2617\pm 0.0029$ \\
\addlinespace[3pt]
\multicolumn{4}{@{}l}{\textbf{CT Scan --- RMSE (target units)}}\\*
LightGBM-Full & 343.8 & $5.10\;[4.40, 5.72]$ & $6.1844\;[4.6835, 7.2040]$ \\*
Endogenous CRAFT-GBM & 319.2 & $3.46\;[3.26, 3.71]$ & $6.2439\;[4.8497, 7.1849]$ \\*
LightGBM-Fixed & 343.8 & $3.99\;[3.58, 4.36]$ & $6.0185\;[4.4152, 6.9385]$ \\*
XGBoost & 343.8 & $7.05\;[6.66, 7.87]$ & $5.9382\;[4.5108, 6.8590]$ \\*
CatBoost & 343.8 & $8.01\;[7.06, 9.18]$ & $6.4751\;[5.1111, 7.2789]$ \\
\addlinespace[3pt]
\multicolumn{4}{@{}l}{\textbf{Sparse 1 --- Binary log loss (nats/observation)}}\\*
LightGBM-Full & 932.2 & $56.75\pm 0.97$ & $0.5041\pm 0.0012$ \\*
Endogenous CRAFT-GBM & 913.2 & $16.58\pm 0.22$ & $0.4991\pm 0.0013$ \\*
LightGBM-Fixed & 932.2 & $36.32\pm 0.52$ & $0.5098\pm 0.0015$ \\*
XGBoost & 932.2 & $108.63\pm 2.28$ & $0.5262\pm 0.0014$ \\*
CatBoost & 932.2 & $96.43\pm 0.99$ & $0.5262\pm 0.0016$ \\
\addlinespace[3pt]
\multicolumn{4}{@{}l}{\textbf{Sparse 2 --- Binary log loss (nats/observation)}}\\*
LightGBM-Full & 615.6 & $2.97\pm 0.11$ & $0.4686\pm 0.0016$ \\*
Endogenous CRAFT-GBM & 579.6 & $2.57\pm 0.07$ & $0.4650\pm 0.0016$ \\*
LightGBM-Fixed & 615.6 & $2.89\pm 0.10$ & $0.4738\pm 0.0021$ \\*
XGBoost & 615.6 & $25.84\pm 0.52$ & $0.4748\pm 0.0023$ \\*
CatBoost & 615.6 & $20.01\pm 0.55$ & $0.4703\pm 0.0021$ \\
\addlinespace[3pt]
\multicolumn{4}{@{}l}{\textbf{Dense 1 --- RMSE (target units)}}\\*
LightGBM-Full & 171.2 & $15.71\pm 0.59$ & $1.0659\pm 0.0029$ \\*
Endogenous CRAFT-GBM & 176.2 & $9.11\pm 0.08$ & $1.0565\pm 0.0039$ \\*
LightGBM-Fixed & 171.2 & $8.97\pm 0.14$ & $1.1664\pm 0.0316$ \\*
XGBoost & 171.2 & $49.87\pm 1.63$ & $1.1620\pm 0.0212$ \\*
CatBoost & 171.2 & $28.10\pm 0.73$ & $1.0913\pm 0.0101$ \\
\addlinespace[3pt]
\multicolumn{4}{@{}l}{\textbf{Dense 2 --- Binary log loss (nats/observation)}}\\*
LightGBM-Full & 1108.2 & $78.67\pm 6.67$ & $0.3860\pm 0.0190$ \\*
Endogenous CRAFT-GBM & 1170.8 & $19.68\pm 1.33$ & $0.3691\pm 0.0205$ \\*
LightGBM-Fixed & 1108.2 & $24.01\pm 2.32$ & $0.3933\pm 0.0185$ \\*
XGBoost & 1108.2 & $199.59\pm 13.30$ & $0.3684\pm 0.0169$ \\*
CatBoost & 1108.2 & $93.92\pm 10.27$ & $0.3733\pm 0.0172$ \\
\addlinespace[3pt]
\multicolumn{4}{@{}l}{\textbf{HAR +4 --- Multiclass log loss (nats/observation)}}\\*
LightGBM-Full & 111.2 & $9.63\pm 0.08$ & $0.1727\pm 0.0039$ \\*
Endogenous CRAFT-GBM & 113.8 & $7.42\pm 0.07$ & $0.1710\pm 0.0010$ \\*
LightGBM-Fixed & 111.2 & $7.56\pm 0.03$ & $0.1665\pm 0.0034$ \\*
XGBoost & 111.2 & $47.96\pm 1.25$ & $0.2082\pm 0.0042$ \\*
CatBoost & 111.2 & $92.62\pm 1.05$ & $0.2599\pm 0.0028$ \\
\addlinespace[3pt]
\multicolumn{4}{@{}l}{\textbf{HAR\_N --- Multiclass log loss (nats/observation)}}\\*
LightGBM-Full & 111.6 & $16.51\pm 0.33$ & $0.1734\pm 0.0031$ \\*
Endogenous CRAFT-GBM & 112.0 & $12.38\pm 0.21$ & $0.1714\pm 0.0034$ \\*
LightGBM-Fixed & 111.6 & $12.85\pm 0.05$ & $0.1708\pm 0.0049$ \\*
XGBoost & 111.6 & $87.46\pm 1.15$ & $0.2085\pm 0.0055$ \\*
CatBoost & 111.6 & $179.86\pm 3.19$ & $0.2638\pm 0.0041$ \\
\addlinespace[3pt]
\multicolumn{4}{@{}l}{\textbf{YearMSD +4 --- RMSE (target units)}}\\*
LightGBM-Full & 326.8 & $19.70\pm 1.06$ & $9.0245\pm 0.0059$ \\*
Endogenous CRAFT-GBM & 322.0 & $10.95\pm 0.47$ & $9.0093\pm 0.0050$ \\*
LightGBM-Fixed & 326.8 & $11.27\pm 0.26$ & $9.0248\pm 0.0041$ \\*
XGBoost & 326.8 & $57.61\pm 2.44$ & $9.0086\pm 0.0078$ \\*
CatBoost & 326.8 & $29.92\pm 1.15$ & $9.1362\pm 0.0109$ \\
\addlinespace[3pt]
\multicolumn{4}{@{}l}{\textbf{YearMSD\_N --- RMSE (target units)}}\\*
LightGBM-Full & 316.8 & $32.38\pm 1.13$ & $9.0337\pm 0.0050$ \\*
Endogenous CRAFT-GBM & 321.8 & $16.38\pm 1.07$ & $9.0129\pm 0.0085$ \\*
LightGBM-Fixed & 316.8 & $16.65\pm 1.01$ & $9.0421\pm 0.0052$ \\*
XGBoost & 316.8 & $111.67\pm 2.20$ & $9.0222\pm 0.0059$ \\*
CatBoost & 316.8 & $47.22\pm 0.88$ & $9.1438\pm 0.0069$ \\
\end{longtable}
\endgroup
\begingroup
\begin{longtable}{@{}lrrrr@{}}
\caption{RQ1: paired percentage changes relative to LightGBM-Full. Entries apply $100(r-1)$ to geometric-mean ratios and pointwise paired 95\% interval endpoints, whereas CT brackets are descriptive fold ranges. Negative values are favorable.}\label{tab:rq1-numerical-paired}\\
\toprule
Model & $\Delta$ time (\%) & Interval & $\Delta$ loss (\%) & Interval \\
\midrule
\endfirsthead
\multicolumn{5}{@{}l}{\tablename~\thetable{} (continued)}\\
\toprule
Model & $\Delta$ time (\%) & Interval & $\Delta$ loss (\%) & Interval \\
\midrule
\endhead
\midrule
\multicolumn{5}{r}{Continued on next page}\\
\endfoot
\bottomrule
\endlastfoot
\addlinespace[3pt]
\multicolumn{5}{@{}l}{\textbf{HIGGS}}\\*
Endogenous CRAFT-GBM & -3.59 & [-7.27, 0.23] & 0.39 & [0.14, 0.64] \\*
LightGBM-Fixed & -7.09 & [-8.05, -6.11] & 0.28 & [0.20, 0.35] \\*
XGBoost & 42.44 & [41.36, 43.53] & 0.39 & [0.27, 0.51] \\*
CatBoost & 48.49 & [48.01, 48.97] & 3.44 & [3.39, 3.50] \\
\addlinespace[3pt]
\multicolumn{5}{@{}l}{\textbf{Epsilon}}\\*
Endogenous CRAFT-GBM & -75.37 & [-75.72, -75.01] & -1.58 & [-1.81, -1.35] \\*
LightGBM-Fixed & -67.41 & [-67.49, -67.33] & 1.31 & [1.16, 1.46] \\*
XGBoost & 255.00 & [254.45, 255.55] & 1.93 & [1.72, 2.15] \\*
CatBoost & -10.49 & [-10.73, -10.24] & 6.00 & [5.87, 6.12] \\
\addlinespace[3pt]
\multicolumn{5}{@{}l}{\textbf{Bosch}}\\*
Endogenous CRAFT-GBM & -49.61 & [-58.50, -38.81] & 0.05 & [-0.40, 0.51] \\*
LightGBM-Fixed & -10.26 & [-10.32, -10.20] & 0.14 & [-0.07, 0.35] \\*
XGBoost & -45.99 & [-46.38, -45.61] & 0.08 & [-0.08, 0.25] \\*
CatBoost & 175.75 & [175.17, 176.34] & 1.04 & [0.89, 1.19] \\
\addlinespace[3pt]
\multicolumn{5}{@{}l}{\textbf{YearMSD}}\\*
Endogenous CRAFT-GBM & -15.55 & [-20.44, -10.36] & 0.41 & [0.35, 0.47] \\*
LightGBM-Fixed & -12.28 & [-12.52, -12.04] & -0.08 & [-0.13, -0.02] \\*
XGBoost & 67.20 & [64.10, 70.37] & 0.00 & [-0.07, 0.06] \\*
CatBoost & 76.86 & [76.36, 77.35] & 1.58 & [1.52, 1.65] \\
\addlinespace[3pt]
\multicolumn{5}{@{}l}{\textbf{APS Failure}}\\*
Endogenous CRAFT-GBM & -21.48 & [-22.91, -20.02] & -3.20 & [-4.85, -1.53] \\*
LightGBM-Fixed & -10.71 & [-11.80, -9.60] & -0.90 & [-3.06, 1.31] \\*
XGBoost & 48.51 & [46.12, 50.93] & -2.65 & [-6.00, 0.82] \\*
CatBoost & 96.18 & [91.19, 101.31] & -5.13 & [-6.46, -3.78] \\
\addlinespace[3pt]
\multicolumn{5}{@{}l}{\textbf{KDD 2009}}\\*
Endogenous CRAFT-GBM & -32.19 & [-33.26, -31.10] & -1.58 & [-3.04, -0.10] \\*
LightGBM-Fixed & -27.45 & [-28.35, -26.55] & -14.23 & [-14.88, -13.58] \\*
XGBoost & 241.64 & [233.78, 249.68] & 35.91 & [35.04, 36.78] \\*
CatBoost & 351.08 & [341.58, 360.77] & -37.67 & [-38.21, -37.14] \\
\addlinespace[3pt]
\multicolumn{5}{@{}l}{\textbf{Gisette}}\\*
Endogenous CRAFT-GBM & -42.17 & [-43.17, -41.15] & -2.05 & [-4.50, 0.45] \\*
LightGBM-Fixed & -39.59 & [-40.01, -39.16] & -9.66 & [-12.49, -6.74] \\*
XGBoost & 92.29 & [91.94, 92.64] & -8.99 & [-11.63, -6.26] \\*
CatBoost & 193.21 & [190.08, 196.38] & 13.13 & [8.51, 17.94] \\
\addlinespace[3pt]
\multicolumn{5}{@{}l}{\textbf{Jannis}}\\*
Endogenous CRAFT-GBM & -12.96 & [-18.02, -7.57] & -3.02 & [-4.15, -1.87] \\*
LightGBM-Fixed & -7.21 & [-8.19, -6.21] & -0.58 & [-1.06, -0.10] \\*
XGBoost & 67.12 & [66.50, 67.75] & -8.27 & [-8.57, -7.97] \\*
CatBoost & 68.66 & [68.09, 69.22] & -11.41 & [-11.70, -11.12] \\
\addlinespace[3pt]
\multicolumn{5}{@{}l}{\textbf{HAR}}\\*
Endogenous CRAFT-GBM & 0.11 & [-1.83, 2.09] & 0.44 & [-2.35, 3.32] \\*
LightGBM-Fixed & -2.72 & [-2.96, -2.48] & -1.19 & [-4.92, 2.69] \\*
XGBoost & 260.01 & [258.58, 261.44] & 29.33 & [26.62, 32.09] \\*
CatBoost & 625.35 & [622.02, 628.69] & 55.16 & [50.84, 59.60] \\
\addlinespace[3pt]
\multicolumn{5}{@{}l}{\textbf{CT Scan --- descriptive fold ranges}}\\*
Endogenous CRAFT-GBM & -31.99 & [-40.37, -25.90] & 1.09 & [-0.26, 3.55] \\*
LightGBM-Fixed & -21.68 & [-23.74, -18.62] & -2.73 & [-9.77, 1.16] \\*
XGBoost & 38.56 & [28.77, 51.24] & -3.88 & [-7.60, -0.63] \\*
CatBoost & 57.10 & [48.12, 60.59] & 5.07 & [1.04, 9.13] \\
\addlinespace[3pt]
\multicolumn{5}{@{}l}{\textbf{Sparse 1}}\\*
Endogenous CRAFT-GBM & -70.78 & [-71.80, -69.72] & -1.01 & [-1.40, -0.61] \\*
LightGBM-Fixed & -36.00 & [-36.23, -35.76] & 1.12 & [0.97, 1.26] \\*
XGBoost & 91.40 & [87.51, 95.37] & 4.38 & [4.25, 4.51] \\*
CatBoost & 69.93 & [68.38, 71.49] & 4.38 & [4.24, 4.51] \\
\addlinespace[3pt]
\multicolumn{5}{@{}l}{\textbf{Sparse 2}}\\*
Endogenous CRAFT-GBM & -13.64 & [-14.95, -12.31] & -0.78 & [-0.85, -0.72] \\*
LightGBM-Fixed & -2.80 & [-3.93, -1.65] & 1.11 & [0.94, 1.27] \\*
XGBoost & 769.83 & [735.33, 805.75] & 1.30 & [1.08, 1.53] \\*
CatBoost & 573.55 & [562.27, 585.03] & 0.36 & [0.16, 0.56] \\
\addlinespace[3pt]
\multicolumn{5}{@{}l}{\textbf{Dense 1}}\\*
Endogenous CRAFT-GBM & -41.96 & [-44.36, -39.46] & -0.87 & [-1.29, -0.46] \\*
LightGBM-Fixed & -42.84 & [-44.46, -41.17] & 9.40 & [5.77, 13.16] \\*
XGBoost & 217.58 & [213.57, 221.65] & 9.00 & [6.78, 11.27] \\*
CatBoost & 78.99 & [76.16, 81.87] & 2.38 & [1.37, 3.41] \\
\addlinespace[3pt]
\multicolumn{5}{@{}l}{\textbf{Dense 2}}\\*
Endogenous CRAFT-GBM & -74.95 & [-77.85, -71.68] & -4.41 & [-5.62, -3.18] \\*
LightGBM-Fixed & -69.50 & [-70.79, -68.16] & 1.91 & [0.91, 2.92] \\*
XGBoost & 153.99 & [147.68, 160.46] & -4.55 & [-5.08, -4.02] \\*
CatBoost & 19.16 & [11.82, 26.99] & -3.28 & [-4.19, -2.36] \\
\addlinespace[3pt]
\multicolumn{5}{@{}l}{\textbf{HAR +4}}\\*
Endogenous CRAFT-GBM & -22.96 & [-24.34, -21.55] & -0.96 & [-3.24, 1.38] \\*
LightGBM-Fixed & -21.46 & [-22.19, -20.73] & -3.56 & [-8.20, 1.32] \\*
XGBoost & 398.04 & [378.67, 418.20] & 20.59 & [16.61, 24.72] \\*
CatBoost & 861.97 & [854.57, 869.43] & 50.53 & [45.02, 56.25] \\
\addlinespace[3pt]
\multicolumn{5}{@{}l}{\textbf{HAR\_N}}\\*
Endogenous CRAFT-GBM & -24.99 & [-27.44, -22.47] & -1.15 & [-2.32, 0.03] \\*
LightGBM-Fixed & -22.14 & [-23.71, -20.55] & -1.54 & [-4.59, 1.61] \\*
XGBoost & 429.84 & [409.06, 451.47] & 20.24 & [15.99, 24.65] \\*
CatBoost & 989.54 & [980.86, 998.29] & 52.13 & [49.53, 54.77] \\
\addlinespace[3pt]
\multicolumn{5}{@{}l}{\textbf{YearMSD +4}}\\*
Endogenous CRAFT-GBM & -44.39 & [-48.91, -39.48] & -0.17 & [-0.24, -0.10] \\*
LightGBM-Fixed & -42.70 & [-47.72, -37.19] & 0.00 & [-0.06, 0.07] \\*
XGBoost & 192.66 & [177.38, 208.78] & -0.18 & [-0.26, -0.09] \\*
CatBoost & 51.99 & [46.87, 57.29] & 1.24 & [1.13, 1.34] \\
\addlinespace[3pt]
\multicolumn{5}{@{}l}{\textbf{YearMSD\_N}}\\*
Endogenous CRAFT-GBM & -49.47 & [-54.55, -43.81] & -0.23 & [-0.31, -0.15] \\*
LightGBM-Fixed & -48.62 & [-53.61, -43.10] & 0.09 & [0.02, 0.17] \\*
XGBoost & 245.01 & [230.90, 259.73] & -0.13 & [-0.20, -0.05] \\*
CatBoost & 45.89 & [40.66, 51.32] & 1.22 & [1.13, 1.30] \\
\end{longtable}
\endgroup
\begingroup
\begin{longtable}{@{}lrr@{}}
\caption{RQ1: synthetic oracle diagnostics used only for evaluation. Values are mean $\pm$ sample SD over five generated problems. The first metric is oracle cross-entropy for Sparse 1/2 and function MSE for Dense 1, while Bernoulli KL is defined only for Sparse 1/2. These quantities do not guide training or selection.}\label{tab:rq1-numerical-oracle}\\
\toprule
Model & Oracle CE / function MSE & Bernoulli KL \\
\midrule
\endfirsthead
\multicolumn{3}{@{}l}{\tablename~\thetable{} (continued)}\\
\toprule
Model & Oracle CE / function MSE & Bernoulli KL \\
\midrule
\endhead
\midrule
\multicolumn{3}{r}{Continued on next page}\\
\endfoot
\bottomrule
\endlastfoot
\addlinespace[3pt]
\multicolumn{3}{@{}l}{\textbf{Sparse 1 --- Oracle cross-entropy and Bernoulli KL (nats/observation)}}\\*
LightGBM-Full & $0.5035\pm 0.0013$ & $0.0312\pm 5.50\!\times\!10^{-4}$ \\*
Endogenous CRAFT-GBM & $0.4984\pm 0.0018$ & $0.0260\pm 0.0013$ \\*
LightGBM-Fixed & $0.5094\pm 0.0014$ & $0.0370\pm 7.04\!\times\!10^{-4}$ \\*
XGBoost & $0.5259\pm 0.0016$ & $0.0535\pm 9.64\!\times\!10^{-4}$ \\*
CatBoost & $0.5258\pm 0.0016$ & $0.0534\pm 0.0011$ \\
\addlinespace[3pt]
\multicolumn{3}{@{}l}{\textbf{Sparse 2 --- Oracle cross-entropy and Bernoulli KL (nats/observation)}}\\*
LightGBM-Full & $0.4681\pm 8.17\!\times\!10^{-4}$ & $0.0135\pm 9.08\!\times\!10^{-4}$ \\*
Endogenous CRAFT-GBM & $0.4646\pm 7.04\!\times\!10^{-4}$ & $0.0099\pm 7.55\!\times\!10^{-4}$ \\*
LightGBM-Fixed & $0.4733\pm 0.0013$ & $0.0187\pm 0.0016$ \\*
XGBoost & $0.4744\pm 0.0016$ & $0.0197\pm 0.0018$ \\*
CatBoost & $0.4699\pm 0.0015$ & $0.0153\pm 0.0017$ \\
\addlinespace[3pt]
\multicolumn{3}{@{}l}{\textbf{Dense 1 --- Function MSE (squared target units)}}\\*
LightGBM-Full & $0.1355\pm 0.0050$ & \textemdash \\*
Endogenous CRAFT-GBM & $0.1162\pm 0.0078$ & \textemdash \\*
LightGBM-Fixed & $0.3633\pm 0.0761$ & \textemdash \\*
XGBoost & $0.3520\pm 0.0439$ & \textemdash \\*
CatBoost & $0.1918\pm 0.0196$ & \textemdash \\
\end{longtable}
\endgroup

\subsection{RQ2: Exogenous Fractions and Frozen Tuning Selections}

Within each campaign, the same LightGBM-Full reference serves all five fractions \(f\)
in a paired comparison. Exogenous CRAFT-GBM uses interval point estimates \(c_{j,t}\)
and its fresh same-campaign Full reference, while the other non-ablated models retain
their original paired reference. The original, updated \craft{}, and ablation
Full references are listed separately in the absolute-outcome tables.
Data, tuning selections, and training settings
are shared across these campaigns. CRAFT-GBM's first two rounds use all features,
whereas its later rounds and LightGBM-Fixed use the same target
\(K_f=\max\{1,\lfloor fp+1/2\rfloor\}\), encoded through each implementation's
native fraction parameter \(f\). XGBoost/CatBoost instead receive \(f\) directly.
RQ2 fixes \(\delta=0.05\) and \(n_{\mathrm{ref}}=1000\).
LightGBM-Full's round count \(T^\star\) and learning rate \(\eta^\star\) are selected on an 80/20 inner
training/validation split, group-disjoint for HAR and CT. All measured
models then refit on the full training set with that selected pair \((T^\star,\eta^\star)\).
The 34 tuning studies use 20 trials each: rounds in \([50,2000]\) and learning
rate in \([0.01,0.20]\), both searched on logarithmic scales. Optuna's seeded
TPE search \citep{akiba2019optuna,bergstra2011algorithms} includes
\((300,0.05)\), uses five startup trials, and selects by
inner-validation loss at the requested endpoint, without early stopping.
The inner-split seed is 20260916, the TPE seed is 20260917, and the tuning
model seed is 20260918. Measured refits use the measurement seeds above.
Tuning uses eight cores and is excluded from measured sixteen-core times.
Selected rates \(\eta^\star\) below are rounded only for display, while fitting uses full precision.

\begingroup
\begin{longtable}{@{}lrrr@{}}
\caption{RQ2: aggregate paired percentage changes. The 14 original families receive equal weight. Brackets are pointwise 95\% family-resampling bands describing benchmark-composition sensitivity.}\label{tab:rq2-numerical-summary}\\
\toprule
Model & $f$ & $\Delta$ time (\%) & $\Delta$ loss (\%) \\
\midrule
\endfirsthead
\multicolumn{4}{@{}l}{\tablename~\thetable{} (continued)}\\
\toprule
Model & $f$ & $\Delta$ time (\%) & $\Delta$ loss (\%) \\
\midrule
\endhead
\midrule
\multicolumn{4}{r}{Continued on next page}\\
\endfoot
\bottomrule
\endlastfoot
Exogenous CRAFT-GBM & 0.1 & -62.74 [-73.74, -47.52] & -0.04 [-2.83, 2.70] \\
Exogenous CRAFT-GBM & 0.3 & -44.20 [-54.21, -32.29] & -1.16 [-2.76, 0.29] \\
Exogenous CRAFT-GBM & 0.5 & -28.02 [-36.05, -19.37] & -0.27 [-1.77, 1.71] \\
Exogenous CRAFT-GBM & 0.7 & -11.77 [-17.23, -6.25] & -0.38 [-1.48, 0.96] \\
Exogenous CRAFT-GBM & 0.9 & -1.81 [-4.02, 0.51] & -0.26 [-0.49, -0.05] \\
LightGBM-Fixed & 0.1 & -49.22 [-60.91, -35.28] & 4.60 [-3.04, 17.28] \\
LightGBM-Fixed & 0.3 & -35.17 [-44.82, -24.58] & -1.44 [-4.11, 1.05] \\
LightGBM-Fixed & 0.5 & -23.25 [-30.28, -15.96] & -1.34 [-2.76, -0.12] \\
\end{longtable}
\endgroup
\begingroup
\begin{longtable}{@{}lrrrrr@{}}
\caption{RQ2: all 34 frozen tuning selections. Each study has 20 valid, unique trials (680 total). $T^\star$ and $\eta^\star$ are transferred to every measured method and fraction. Boundary marks a selected search limit, while tuning wall time is separate from measured training time.}\label{tab:rq2-numerical-tuning}\\
\toprule
Dataset / identity & $T^\star$ & $\eta^\star$ & Validation loss & Boundary & Tuning (s) \\
\midrule
\endfirsthead
\multicolumn{6}{@{}l}{\tablename~\thetable{} (continued)}\\
\toprule
Dataset / identity & $T^\star$ & $\eta^\star$ & Validation loss & Boundary & Tuning (s) \\
\midrule
\endhead
\midrule
\multicolumn{6}{r}{Continued on next page}\\
\endfoot
\bottomrule
\endlastfoot
HIGGS & 1955 & 0.152296 & 0.48986 & none & 4504.9 \\
Epsilon & 1842 & 0.066851 & 0.27077 & none & 3556.7 \\
Bosch & 1780 & 0.013433 & 0.03087 & none & 2448.3 \\
YearMSD & 1865 & 0.065079 & 8.72367 & none & 436.4 \\
APS Failure & 571 & 0.023038 & 0.01752 & none & 146.5 \\
KDD 2009 & 93 & 0.022992 & 0.07730 & none & 116.2 \\
Gisette & 1652 & 0.010419 & 0.06000 & none & 668.9 \\
Jannis & 862 & 0.013581 & 0.70006 & none & 348.4 \\
\end{longtable}
\endgroup
\begingroup
\begin{longtable}{@{}>{\raggedright\arraybackslash}p{0.23\linewidth}rrrr@{}}
\caption{RQ2: absolute outcomes for every measured configuration. Each row summarizes five replicates: mean $\pm$ sample SD, except CT, which uses mean [minimum, maximum]. Full (CRAFT-GBM ref.) is the reference for Exogenous CRAFT-GBM, Full (ablation ref.) serves Mean and Exposure ablations, and the unqualified Full row serves the remaining models. Each reference was measured in its own campaign.}\label{tab:rq2-numerical-raw}\\
\toprule
Model & $f$ & Mean rounds & Training (s) & Test loss \\
\midrule
\endfirsthead
\multicolumn{5}{@{}l}{\tablename~\thetable{} (continued)}\\
\toprule
Model & $f$ & Mean rounds & Training (s) & Test loss \\
\midrule
\endhead
\midrule
\multicolumn{5}{r}{Continued on next page}\\
\endfoot
\bottomrule
\endlastfoot
\addlinespace[3pt]
\multicolumn{5}{@{}l}{\textbf{HIGGS --- Binary log loss (nats/observation)}}\\*
LightGBM-Full & \textemdash & 1955.0 & $282.24\pm 0.48$ & $0.4873\pm 1.45\!\times\!10^{-4}$ \\*
LightGBM-Full (CRAFT-GBM ref.) & \textemdash & 1955.0 & $281.97\pm 0.50$ & $0.4873\pm 1.45\!\times\!10^{-4}$ \\*
LightGBM-Full (ablation ref.) & \textemdash & 1955.0 & $281.05\pm 0.28$ & $0.4873\pm 1.45\!\times\!10^{-4}$ \\*
Exogenous CRAFT-GBM & 0.1 & 1955.0 & $219.95\pm 1.48$ & $0.5343\pm 0.0028$ \\*
Exogenous CRAFT-GBM & 0.3 & 1955.0 & $232.85\pm 0.43$ & $0.5071\pm 0.0022$ \\*
Exogenous CRAFT-GBM & 0.5 & 1955.0 & $243.14\pm 1.09$ & $0.5000\pm 3.32\!\times\!10^{-4}$ \\*
Exogenous CRAFT-GBM & 0.7 & 1955.0 & $266.25\pm 0.88$ & $0.4903\pm 4.46\!\times\!10^{-4}$ \\
\end{longtable}
\endgroup
\begingroup
\begin{longtable}{@{}>{\raggedright\arraybackslash}p{0.23\linewidth}rrrrr@{}}
\caption{RQ2: paired percentage changes relative to LightGBM-Full. Entries apply $100(r-1)$ to geometric-mean ratios and pointwise paired 95\% interval endpoints, whereas CT brackets are descriptive fold ranges. Negative values are favorable. Exogenous CRAFT-GBM and the ablations each use their same-campaign Full reference, while the remaining models use the original Full reference.}\label{tab:rq2-numerical-paired}\\
\toprule
Model & $f$ & $\Delta$ time (\%) & Interval & $\Delta$ loss (\%) & Interval \\
\midrule
\endfirsthead
\multicolumn{6}{@{}l}{\tablename~\thetable{} (continued)}\\
\toprule
Model & $f$ & $\Delta$ time (\%) & Interval & $\Delta$ loss (\%) & Interval \\
\midrule
\endhead
\midrule
\multicolumn{6}{r}{Continued on next page}\\
\endfoot
\bottomrule
\endlastfoot
\addlinespace[3pt]
\multicolumn{6}{@{}l}{\textbf{HIGGS}}\\*
Exogenous CRAFT-GBM & 0.1 & -22.00 & [-22.72, -21.27] & 9.66 & [8.94, 10.38] \\*
Exogenous CRAFT-GBM & 0.3 & -17.42 & [-17.73, -17.11] & 4.07 & [3.53, 4.62] \\*
Exogenous CRAFT-GBM & 0.5 & -13.77 & [-14.36, -13.18] & 2.60 & [2.48, 2.72] \\*
Exogenous CRAFT-GBM & 0.7 & -5.57 & [-5.98, -5.17] & 0.63 & [0.49, 0.76] \\*
Exogenous CRAFT-GBM & 0.9 & -2.83 & [-3.09, -2.57] & 0.49 & [0.34, 0.65] \\*
LightGBM-Fixed & 0.1 & -14.81 & [-15.38, -14.23] & 5.82 & [5.72, 5.92] \\*
LightGBM-Fixed & 0.3 & -11.68 & [-12.54, -10.82] & 1.89 & [1.69, 2.10] \\
\end{longtable}
\endgroup
\begingroup
\begin{longtable}{@{}lrrr@{}}
\caption{RQ2: synthetic oracle diagnostics used only for evaluation. Values are mean $\pm$ sample SD over five generated problems. The first metric is oracle cross-entropy for Sparse 1/2 and function MSE for Dense 1, while Bernoulli KL is defined only for Sparse 1/2. These quantities do not guide training or selection. Full (CRAFT-GBM ref.) identifies the fresh Exogenous CRAFT-GBM reference. Full (ablation ref.) identifies the reference shared by Mean and Exposure ablations.}\label{tab:rq2-numerical-oracle}\\
\toprule
Model & $f$ & Oracle CE / function MSE & Bernoulli KL \\
\midrule
\endfirsthead
\multicolumn{4}{@{}l}{\tablename~\thetable{} (continued)}\\
\toprule
Model & $f$ & Oracle CE / function MSE & Bernoulli KL \\
\midrule
\endhead
\midrule
\multicolumn{4}{r}{Continued on next page}\\
\endfoot
\bottomrule
\endlastfoot
\addlinespace[3pt]
\multicolumn{4}{@{}l}{\textbf{Sparse 1 --- Oracle cross-entropy and Bernoulli KL (nats/observation)}}\\*
LightGBM-Full (CRAFT-GBM ref.) & \textemdash & $0.4969\pm 0.0011$ & $0.0245\pm 0.0010$ \\*
LightGBM-Full (ablation ref.) & \textemdash & $0.4969\pm 0.0011$ & $0.0245\pm 0.0010$ \\*
LightGBM-Full & \textemdash & $0.4969\pm 0.0011$ & $0.0245\pm 0.0010$ \\*
Exogenous CRAFT-GBM & 0.1 & $0.4889\pm 0.0014$ & $0.0165\pm 8.63\!\times\!10^{-4}$ \\*
Exogenous CRAFT-GBM & 0.3 & $0.4932\pm 0.0012$ & $0.0208\pm 7.72\!\times\!10^{-4}$ \\*
Exogenous CRAFT-GBM & 0.5 & $0.4951\pm 9.26\!\times\!10^{-4}$ & $0.0227\pm 7.03\!\times\!10^{-4}$ \\*
Exogenous CRAFT-GBM & 0.7 & $0.4959\pm 9.45\!\times\!10^{-4}$ & $0.0235\pm 8.55\!\times\!10^{-4}$ \\
\end{longtable}
\endgroup

\clearpage

\section{When Can Dynamic Feature Screening Reduce Work?}
\label{sec:complexity}

Forming all bounded endpoints \(L_{j,t},U_{j,t}\) from cached boundary values \(u_\rho(v^+;\alpha)\) costs \(O(p)\).
New boundary-cache entries require the numerical root calculation described
in Appendix~\ref{app:eb-boundary}. The default rule for a set of possible-best features \(S_t\) then finds
\(C_t=\max_jL_{j,t}\) and retains every \(U_{j,t}\geq C_t\) in two linear
scans, without sorting. In Exogenous \(K\) mode, the implementation applies
\texttt{std::nth\_element} with the total point-estimate \(c_{j,t}\) ordering, then sorts the
\(K_f\) retained internal positions for the downstream mask. Its expected
selection cost is \(O(p+K_f\log K_f)\), with \(O(p)\) reusable scratch. This
is not a worst-case linear-time claim. Boundary numerators \(u_\rho(v^+;\alpha)\) are memoized on
the upward-rounded squared-error grid \(v^+\), shared across features, so each occupied
grid point requires a root calculation only once. The current observation
count \(N_{j,t-1}\) remains the radius \(r_{j,t-1}\) denominator.

Before each output tree, the implementation clears and collects the root-gain \(G_{j,m,t}\)
slots for the \(K_t=|S_t|\) selected features. Collecting and transforming
these gains costs \(O(MK_t)\). When the selection \(S_t\) changes, the incremental
mask update costs \(O(K_{t-1}+K_t)\), whereas an unchanged set requires no mask or
list rebuild. One-time feature initialization costs \(O(p\log p)\), including
the sort of canonical feature IDs. The gain
histories, endpoint scratch, and top-\(K\) scratch require \(O(p)\) storage,
and the shared boundary cache requires
\(O(128[1+\log(1+V_{\max})])\) entries, where
\(V_{\max}=\max_j V_{j,t}\). Updating the squared-error sums \(V_{j,t}\) costs
\(O(K_t)\) per completed round. The selected-count history additionally stores
one value per completed round, requiring \(O(t)\) space.
The root-gain buffer and fixed feature mappings
require \(O(p)\), in addition to native Dataset storage. \craft{} stores no
modified or rebuilt EFB representation.

Let \(e_a\) be method \(a\)'s realized completed-round endpoint,
\(F_{a,t}\) its feature-dependent histogram and split-search work, and
\(m_{a,t}\) its training-loss evaluation and endpoint-controller work.
Let \(b_{a,t}\) include remaining native per-round work, such as gradient and
Hessian calculation, row sampling, and prediction updates.
For \craft{}, let \(a_{a,t}\geq0\) be mapping, interval, selection, masking,
and gain-collection work, which is zero for native LightGBM. If \(C_a^{(0)}\) contains
dataset loading, initial-loss evaluation, model materialization, and other
method-specific endpoint costs, the honest decomposition is
\begin{equation}
C_a=C_a^{(0)}+
\sum_{t=1}^{e_a}\bigl(F_{a,t}+b_{a,t}+m_{a,t}+a_{a,t}\bigr),
\qquad
a_{\mathrm{LGBM},t}=0.
\label{eq:work-identity}
\end{equation}
In RQ1, the endogenous endpoint is \(e_a=\tau_a\), whereas fixed-horizon
comparators use the transferred pre-run endpoint \(t^{(0)}\). At a common fixed endpoint,
\craft{} is faster only when its saved histogram and split-search work \(F_{a,t}\) exceeds
its added selector work \(a_{a,t}\) and the net change in other native, metric/controller,
and endpoint costs \(C_a^{(0)}\). Under adaptive stopping, tail
length also enters. The measured runtime consequently includes the entire
executed prefix, any patience tail, training-loss \(L_t\) evaluations, the controller,
and model serialization. Final-test prediction is outside this timing scope.

For physical accounting, let \(\mathcal H_{a,t}\subseteq\mathcal E\) be the
EFB groups actually visited and
\(\mathcal J_{a,t}\subseteq\{1,\ldots,p\}\) the features actually considered
as split candidates in arm \(a\in\{\mathrm L,\mathrm J\}\). One may
decompose feature-dependent native work as
\[
F_{a,t}
=\sum_{b\in\mathcal H_{a,t}}g^a_{b,t}
+\sum_{j\in\mathcal J_{a,t}}e^a_{j,t},
\]
where the first term includes shared packed-storage and histogram work and the
second includes feature-specific split evaluation. Native LightGBM has
\(\mathcal J_{\mathrm L,t}\) equal to its native candidate set, and to
\(\{1,\ldots,p\}\) with all native features eligible. A conforming \craft{}
implementation has
\[
\mathcal J_{\mathrm J,t}=S_t.
\]
Its directly implicated
groups, and exactly the groups invoked in the validated forced-column-wise
scope, are \(\mathcal H_{\mathrm J,t}=\mathcal E(S_t)\). Unchanged native
heuristics may still process bins of unselected features inside an implicated
group. Physical costs differ with group membership, density, bin width,
feature fraction, candidate count, memory traffic, and native
sharing. Thus neither the selection size \(K_t\) nor the feature fraction \(K_t/p\), nor even
\(|\mathcal E(S_t)|/B\), is by itself a physical-work guarantee. If \(p=1\),
both models make the sole feature eligible while \craft{} still pays controller
overhead, ruling out a universal no-slowdown theorem.

\end{document}